\pdfoutput=1

\documentclass[11pt]{article}

\usepackage[final]{acl}

\usepackage{times}
\usepackage{latexsym}

\usepackage[T1]{fontenc}
\newcommand{\LESA}{{\bf LESA }}

\usepackage[utf8]{inputenc}

\usepackage{microtype}

\usepackage{inconsolata}

\usepackage{graphicx}

\usepackage{enumitem}
\usepackage{multirow}
\usepackage{makecell}
\usepackage{colortbl}
\usepackage[most]{tcolorbox}
\tcbuselibrary{breakable}
\usepackage[linesnumbered,ruled,vlined,algo2e]{algorithm2e}

\usepackage{booktabs}
\usepackage{amssymb}
\usepackage{amsmath}
\usepackage{soul}
\usepackage{xcolor}
\definecolor{myblue0}{RGB}{192, 214, 234}
\definecolor{myblue1}{RGB}{166, 199, 226}
\definecolor{mygray}{RGB}{224, 224, 232}
\definecolor{mydarkgreen}{RGB}{59, 139, 103}

\title{LESA: Learnable LLM Layer Scaling-Up}

\author{Yifei Yang$^{1,2,3}$, Zouying Cao$^{1,2,3}$, Xinbei Ma$^{1,2,3}$, Yao Yao$^{1,2,3}$, \\ \textbf{Libo Qin$^{4}$, Zhi Chen$^{5}$, Hai Zhao$^{1,2,3}$}\thanks{\,\, Corresponding author. 
} \\
$^1$Department of Computer Science and Engineering, Shanghai Jiao Tong University\\
$^2$Key Laboratory of Shanghai Education Commission for Intelligent Interaction \\
and Cognitive Engineering, Shanghai Jiao Tong University \\
$^3$Shanghai Key Laboratory of Trusted Data Circulation and Governance in Web3 \\
$^4$School of Computer Science and Engineering, Central South University \\
$^5$ByteDance\\
\texttt{yifeiyang@sjtu.edu.cn, zhaohai@cs.sjtu.edu.cn} \\}

\begin{document}
\maketitle
\begin{abstract}
Training Large Language Models (LLMs) from scratch requires immense computational resources, making it prohibitively expensive. Model scaling-up offers a promising solution by leveraging the parameters of smaller models to create larger ones. However, existing depth scaling-up methods rely on empirical heuristic rules for layer duplication, which result in poorer initialization and slower convergence during continual pre-training. We propose \textbf{LESA}, a novel learnable method for depth scaling-up. By concatenating parameters from each layer and applying Singular Value Decomposition, we uncover latent patterns between layers, suggesting that inter-layer parameters can be learned. LESA uses a neural network to predict the parameters inserted between adjacent layers, enabling better initialization and faster training. Experiments show that LESA outperforms existing baselines, achieving superior performance with less than half the computational cost during continual pre-training. Extensive analyses demonstrate its effectiveness across different model sizes and tasks.\footnote{\url{https://github.com/yangyifei729/LESA}}
\end{abstract}

\section{Introduction}
Recent advancements in Natural Language Processing (NLP) have been largely driven by Transformer-based architectures~\cite{vaswani2017attention}, with Large Language Models (LLMs) demonstrating exceptional capabilities in addressing a wide range of complex tasks~\cite{brown2020language, achiam2023gpt, bai2023qwen, touvron2023llama, yang2024qwen2, llama3modelcard, jiang2023mistral, almazrouei2023falcon,bi2024deepseek}. As the parameter size continues to grow, in accordance with scaling laws~\cite{kaplan2020scaling}, the computational resources required to train LLMs from scratch have become increasingly prohibitive, demanding millions of GPU hours and significant energy consumption. This immense resource demand largely arises from the need to randomly reinitialize model parameters, preventing the transfer of ability from existing LLMs.

To address this limitation, a common approach is model scaling-up, which leverages the parameters of smaller models to construct larger ones, either for immediate deployment or as a better initial checkpoint for more effective further continual pre-training. Existing model scaling-up methods can be divided into width scaling-up and depth scaling-up. Width scaling-up~\cite{chen2015net2net, chen2021bert2bert, wang2023learning, samragh2024scaling} primarily involves expanding matrix dimensions, rather than increasing the number of layers~\footnote{A ``layer'' refers to a Transformer block for simplicity.}. In contrast, depth scaling-up involves repurposing trained Transformer blocks from a smaller model to build a larger one with additional layers~\cite{wu2024llama, kim2023solar, gong2019efficient, pan2024preparing, agarwal2024stacking, parmar2024reuse}. This strategy is widely applicable to modern LLMs based on the Transformer architecture, preserving the internal structure, such as matrix sizes. It is also compatible with existing parallel training frameworks, better preserving the model's knowledge, contributing to its increasing popularity in recent model scaling-up approaches.

\begin{figure*}[!tp]
    \centering
    \includegraphics[width=0.98\linewidth,scale=1.00]{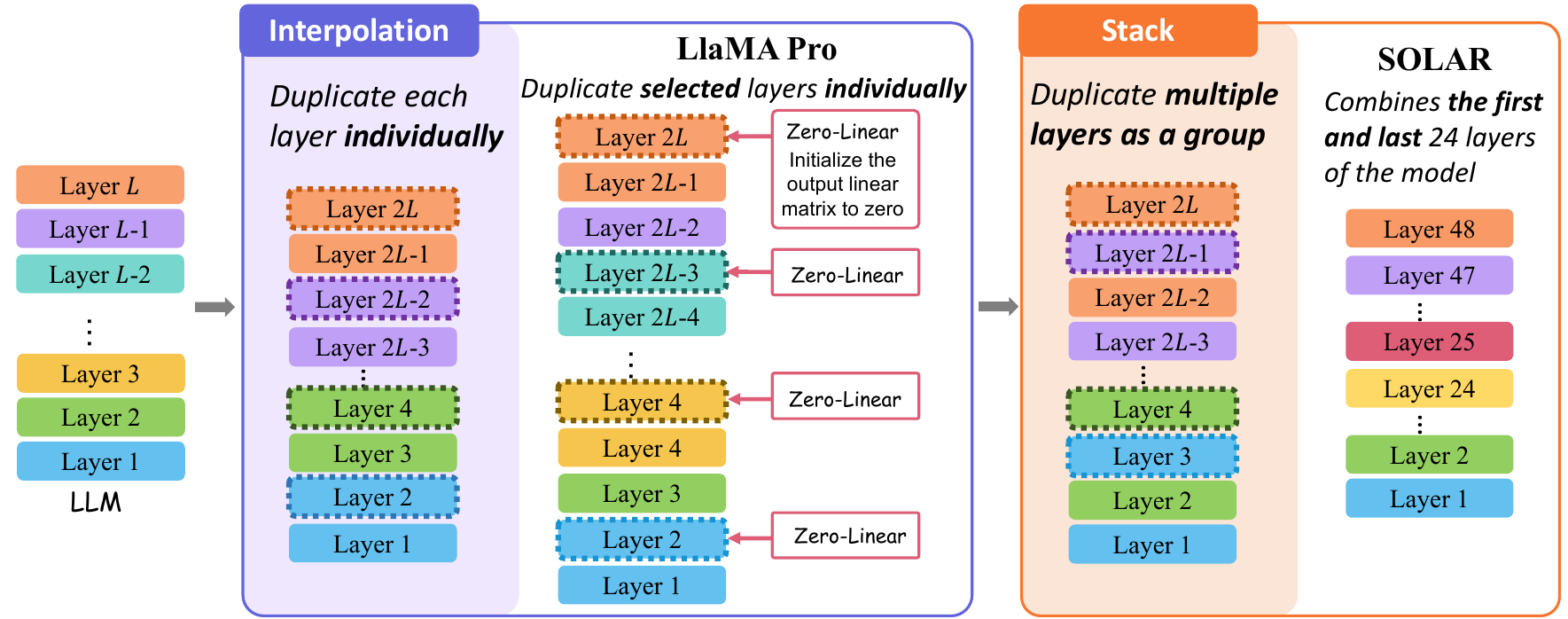}
    \caption{Existing depth scaling-up methods can be categorized into two types: ``Interpolation'' and ``Stack''. LLaMA Pro and SOLAR can be seen as specific examples of these two types. Layers with the same color represent identical parameters, and the dashed boxes indicate those obtained through duplication.}
    \label{fig:related}
\end{figure*}

However, current depth scaling-up methods rely on heuristic rules, typically duplicating one or more blocks before integrating them into the model. These approaches overlook parameter change patterns between layers, limiting the model's ability to specialize each layer effectively. As a result, newly upscaled layers replicate the previous ones, neglecting layer-specific specialization~\cite{voita2019analyzing, voita2019bottom}. This not only leads to suboptimal model initialization performance but also prevents the model from fully utilizing its expanded capacity. By treating all layers equally, these methods fail to capture the nuanced relationships between layers, causing slower convergence during training and yielding less effective models.

In this paper, we propose a novel approach for depth scaling-up called \textbf{LESA} (\textbf{LE}arnable LLM Layer \textbf{S}c\textbf{A}ling-Up). We are the first to observe that by concatenating the parameters of each Transformer block and applying Singular Value Decomposition (SVD), patterns such as continuity can be identified between layers. Based on this observation, it is hypothesized that latent patterns exist between Transformer layers in a well-trained LLM, suggesting that model parameters can be learned across layers. To predict these parameters, we propose training a neural network. Once trained, the network can generate intermediate layers between adjacent layers, insert them into the model for depth scaling-up, and serve as a better initialization checkpoint, enabling faster convergence during continual pre-training. Our key contributions are summarized as:
\begin{itemize}
\item We first observe, through SVD, latent patterns such as continuity between Transformer layers, suggesting that inter-layer parameters can potentially be learned.
\item We introduce \textbf{LESA}, which predicts intermediate layer parameters from adjacent layers for depth scaling-up. Experiments show that \textbf{LESA} outperforms existing baselines, with better model initialization and faster convergence during continual pre-training.
\item Extensive experiments confirm that \textbf{LESA} works across various model sizes and families, including domain-specific tasks like code-related tasks. We also perform ablation studies to explore different method configurations.
\end{itemize}

\section{Related Works}
\subsection{Model Scaling-up}
Model scaling-up can be broadly categorized into width and depth scaling-up. Width scaling-up increases the matrix size while ensuring that the output of a layer or consecutive layers remains consistent with the output of the original network before expansion. Net2Net~\cite{chen2015net2net} is one of the first to transfer parameters from a smaller model to initialize a larger one using function-preserving transformations. bert2BERT~\cite{chen2021bert2bert} extends this approach to Transformer-based models. LiGO~\cite{wang2023learning} learns a linear mapping to initialize larger models. HyperCloning~\cite{samragh2024scaling} expands LLM to fit a larger model with more hidden dimensions. However, while these methods increase matrix size, they are less compatible with parallel training frameworks, which are better suited for depth scaling-up. Moreover, depth scaling-up better preserves the model's knowledge.

Current depth scaling-up methods expand the model by duplicating and adding layers based on heuristic rules, which can be broadly categorized into "Interpolation" and "Stack"~\cite{pan2024preparing}, as shown in Figure~\ref{fig:related}. Interpolation involves adding a copy of each layer after the original, while Stack treats consecutive layers as a group and duplicates them together. Recent popular methods like LLaMA Pro~\cite{wu2024llama} and SOLAR~\cite{kim2023solar} can be seen as special cases of these two types. LLaMA Pro copies only a selected few layers, while SOLAR duplicates the first 24 and the last 24 layers of a previous 32-layer model and combines them. However, these methods are based on heuristic rules, which hinder layer specialization, leading to suboptimal performance and limiting the model's potential.

\subsection{Progressive Training}
Progressive training involves gradually transitioning from simpler, smaller models to more complex, larger ones~\cite{chang2017multi, wen2020autogrow, dong2020towards, wei2016network,fayek2020progressive}. It is often combined with model scaling-up, where the model size is progressively increased during training. Prior to the era of LLMs, many methods~\cite{chen2021bert2bert, gu2020transformer, wang2023learning, yang2020progressively, yao2023masked} are developed to train smaller models, such as BERT~\cite{devlin2018bert}. In recent years, LLaMA Pro~\cite{wu2024llama} and Apollo~\cite{pan2024preparing} have applied progressive learning and model scaling-up strategies to train LLMs. YODA~\cite{lu2024yoda} introduces a novel teacher-student progressive learning framework that enhances model fine-tuning by emulating the teacher-student educational process. \citeauthor{Du2024Stacking} offer a comprehensive evaluation and empirical guidelines for progressive learning and model scaling-up.

\section{Method}
This section discusses the patterns observed between model layers through SVD analysis of the model's parameters. Based on these patterns, we hypothesize that there are underlying patterns in the trained model that can be learned by a neural network. We then use this trained network to predict intermediate layers that can be inserted between adjacent layers for depth scaling-up.

\subsection{SVD-Based Layer Pattern}\label{sec:pattern}
Inspired by recent work using SVD for LLM compression or merging~\cite{wang2024svd,stoica2024model,wang2024basis}, it is realized that SVD can map the model's parameters into one space for analysis. Specifically, assume we have weight matrices $\mathcal{W}_1, \mathcal{W}_2, \dots, \mathcal{W}_L$ from $L$ layers of an LLM, where $\mathcal{W}_i \in \mathbb{R}^{d1 \times d2}$ represents a matrix from each Transformer block, such as the up-projection matrix in MLP, Query matrix in self-attention. These $L$ matrices can be concatenated horizontally into a single matrix, denoted as denoted as $\mathcal{W} \in \mathbb{R}^{d_1 \times L d_2}$. SVD can be used to decompose this matrix into three components: $U, \Sigma$ and $V^T$.

According to SVD, $\Sigma$ is a diagonal matrix of size $d_1 \times L d_2$, containing the singular values of $\mathcal{W}$. $U$ is a unitary matrix that spans a set of standard orthogonal bases. If we treat $\Sigma$ as a scaling transformation on each orthogonal basis in $U$, then $U\Sigma$ forms a new set of orthogonal bases. For the $i$-th layer's $\mathcal{W}_i$, it can be recovered as $\mathcal{W}_i = U \Sigma V_i$, where $V_i = V^T_{:,(i-1)*d2:i*d2} \in \mathbb{R}^{d_1 \times d_2}$. This means that the parameter $\mathcal{W}_i$ of each layer is a linear combination of the orthogonal bases from $U\Sigma$, with $V_i$ representing the coefficients of this combination. By projecting the parameters of each layer into the space spanned by $U\Sigma$, we can analyze the patterns in the coefficients $V_i$ for the $i$-th layer.

\begin{figure}[!tp]
    \centering
    \includegraphics[width=0.9\linewidth,scale=1.00]{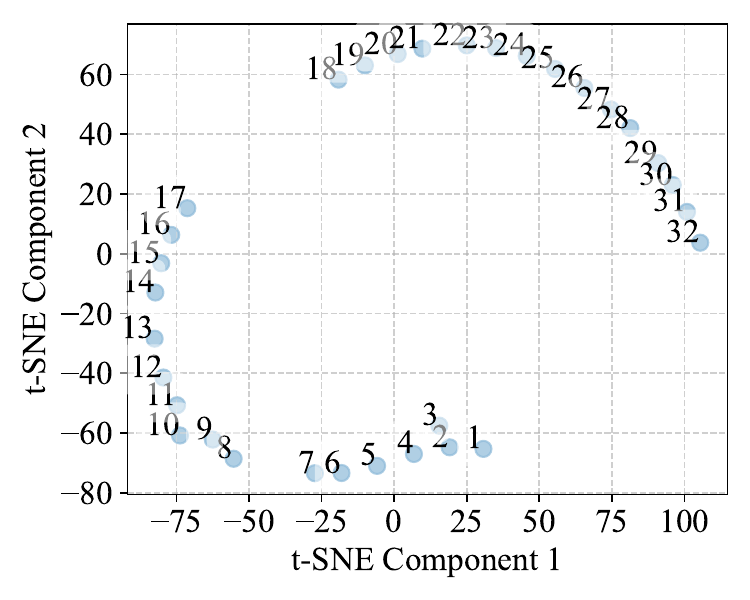}
    \caption{The inter-layer continuity pattern exhibited by the gate\_proj matrix of Llama3-8B in the SVD space. The numbers represent the layer indices.}
    \label{fig:gate_pattern}
\end{figure}

\begin{figure*}[!tp]
    \centering
    \includegraphics[width=0.98\linewidth,scale=1.00]{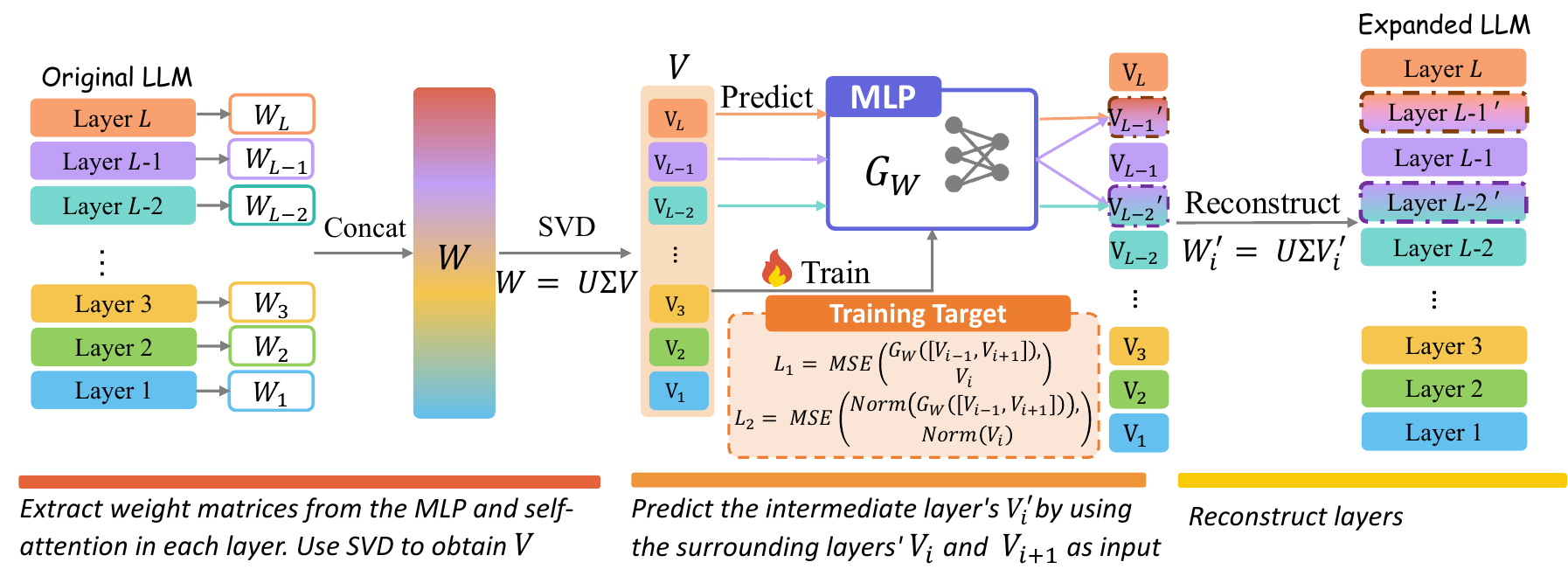}
    \caption{Overview of the proposed \LESA. We first extract the weight matrices from the MLP and self-attention layers. Next, we apply SVD and train a neural network to predict the intermediate layers. Finally, we reconstruct the expanded LLM.}
    \label{fig:main_method}
\end{figure*}

Since larger singular values correspond to eigenvectors that capture more information about the matrix, we select the eigenvector corresponding to the largest singular value (top-1) from each layer's $V_i$ for visualization. We use t-SNE~\cite{van2008visualizing} to reduce $V_i$ to two dimensions. The visualization results of the gate-projection in the MLP of Llama3-8B~\cite{llama3modelcard} are presented in Figure~\ref{fig:gate_pattern}, where we observe a clear continuity in the distribution of these $V_i$. This continuity pattern, derived from the top-1 singular value of the gate-projection using t-SNE, is also present in Llama2~\cite{touvron2023llama2}, Llama3~\cite{llama3modelcard}, and Qwen2~\cite{qwen2}. This suggests that the model's parameters may exhibit unique inter-layer patterns. More visualization results can be found in Appendix~\ref{app:svd_pattern}.

Although this continuity is currently only observed in the gate projection and not in other parameters such as the up-projection or down-projection in the MLP through our t-SNE visualization, this may be due to the limitations of our analysis method. An intuitive approach, therefore, is to use a neural network to learn these potential patterns.

\subsection{Learnable LLM Layer Scaling-Up}\label{sec:lesa}
Inspired by the aforementioned SVD-Based Layer Patterns, we hypothesize that there may be inter-layer patterns in the parameters. However, these patterns might not be easily observed using simple visualization techniques or fitted with specific distributions, such as Gaussian mixtures. Therefore, a direct approach is to learn these patterns through a neural network.

We present our method in Figure~\ref{fig:main_method}. After obtaining $V_i$ as described in Section~\ref{sec:pattern}, we train an MLP $\mathcal{G_W}$ to learn the patterns. Our training objective is to enable the MLP to predict an intermediate layer given any two layers that are one layer apart.

Formally, for a weight matrix $\mathcal{W} \in \mathbf{\left\{
\begin{array}{ll}
\text{q\_proj},\text{k\_proj},\text{v\_proj},\\
\text{o\_proj},\text{up\_proj},\text{down\_proj},\text{gate\_proj}
\end{array}
\right\}}$, we use SVD to obtain $V_i$ following Section~\ref{sec:pattern}. We then train an MLP $\mathcal{G_W}$ specific to $\mathcal{W}$ with the objective of predicting $V_i$ by using the concatenation of $V_{i-1}$ and $V_{i+1}$ as input. We optimize $\mathcal{G_W}$ using MSELoss (Mean Squared Error Loss):
\begin{equation}\label{eq: loss1}
\mathcal{L}_1= MSE(\mathcal{G_W}([V_{i-1}, V_{i+1}]), V_i)
\end{equation}
whose goal is to enable $\mathcal{G_W}$ to predict accurately.

In subsequent experiments, we find that directly training with $\mathcal{L}_1$ will result in the norm of the predicted $\mathcal{G_W}([V_{i-1}, V_{i+1}])$ approaching zero, meaning that the predicted $V_i^{'}$ parameters are close to zero, which leads to parameter degradation. To address this issue, we add a norm loss:
\begin{equation}\label{eq: loss2}
\mathcal{L}_2 = MSE(Norm(\mathcal{G_W}([V_{i-1}, V_{i+1}])), Norm(V_i))
\end{equation}
where the $Norm$ represents the L2 norm, and $\mathcal{L}_2$ aims to ensure that the norm of the model's predicted $V_i^{'}$ is close to that of $V_i$. Thus, the final loss for training $\mathcal{G_W}$ is:
\begin{equation}\label{eq: final}
\mathcal{L} = (1-\lambda)\mathcal{L}_1 + \lambda\mathcal{L}_2
\end{equation}
where $\lambda$ is a hyper-parameter.

Once trained, $\mathcal{G_W}$ can predict the parameters of an intermediate layer based on its surrounding ones. Thus, for adjacent $V_i$ and $V_{i+1}$, we use $\mathcal{G_W}$ to predict the intermediate layer $V_i^{'}$ to insert between them. We then reconstruct $V_i^{'}$ using the $U\Sigma$ decomposition from the previous step, forming the predicted matrix $\mathcal{W}^{'}$, and insert it between the layers to expand the LLM.

\section{Main Experiments}
\subsection{Settings}
\subsubsection{LESA Settings}\label{sec:settings}
We conduct experiments on the Llama3-8B model, which has 32 layers. To construct the training data for $\mathcal{G_W}$, we use consecutive triplets of layers, namely (1, 2, 3), (2, 3, 4), (3, 4, 5), ..., (30, 31, 32), resulting in 30 samples. We define $\mathcal{G_W}$ as a three-layer MLP with a ReLU~\cite{agarap2018deep} activation function, where the hidden dimension is 256. $\mathcal{G_W}$ is trained for 5 epochs on these samples using the AdamW~\cite{loshchilov2017decoupled} optimizer with a learning rate of 1e-3. The $\lambda$ is set to 5e-5. To compare with baselines~\cite{wu2024llama,kim2023solar}, we use \textbf{LESA} to scale up Llama3-8B to 48 layers by inserting an intermediate layer between each pair of adjacent layers in the original 15th to 31st layers. The expanded models all have 11.5 billion parameters.

\subsubsection{Continual Training}
For the models expanded using \LESA and baseline methods, we continue pre-training with Wikipedia data from November 2024, which is released after the training of Llama3-8B and has not been used in its original training. We use the Llama-Factory~\cite{zheng2024llamafactory} training framework, with a cutoff length of 4096, a warmup ratio of 0.1, and a cosine learning rate scheduler. The optimizer is AdamW with a learning rate of 5e-5. The batch size per GPU is 2, with 4 gradient accumulation steps. For LESA and LLaMA Pro, we only train the newly expanded layers, freezing the other layers. Following the original setting, we perform full parameter fine-tuning for SOLAR.

For the Supervised Fine-Tuning (SFT) stage, we use Alpaca-GPT4~\cite{peng2023instruction} for training, following SOLAR. The hyper-parameters are the same as those in the continual pre-training, except that we perform full parameter fine-tuning without freezing any layers for all models. All experiments are conducted on a server with 8 Nvidia A100 80GB GPUs.

\subsection{Benchmarks}
For the continual pre-training models, since they lack instruction-following capabilities, we use the OpenCompass framework \cite{2023opencompass} with the PPL (perplexity)~\footnote{\url{https://opencompass.readthedocs.io/en/latest/get_started/faq.html}} mode for evaluation, focusing on five areas: Reasoning, Language, Knowledge, Examination, and Understanding, with selected benchmarks for each category. \textbf{Reasoning}: CMNLI \cite{xu2020clue}, HellaSwag (HeSw) \cite{zellers2019hellaswag}, PIQA \cite{bisk2019piqa}. \textbf{Language}: CHID \cite{zheng-etal-2019-chid}, WinoGrande (Wino) \cite{sakaguchi2019winogrande}. \textbf{Knowledge}: CommonSenseQA (CSQA) \cite{talmor2018commonsenseqa}, BoolQ \cite{clark2019boolq}. \textbf{Examination}: MMLU \cite{hendryckstest2021}, CMMLU \cite{li2023cmmlu}. \textbf{Understanding}: Race-High/Middle (H/M) \cite{lai2017race}. Evaluations use OpenCompass official scripts in zero-shot or few-shot settings. Scores are computed by OpenCompass, with higher values indicating better performance. We also evaluate the trained models' perplexity on 500 unseen Wikipedia plain sentences.

\begin{figure}[!tp]
    \centering    \includegraphics[width=0.98\linewidth,scale=1.00]{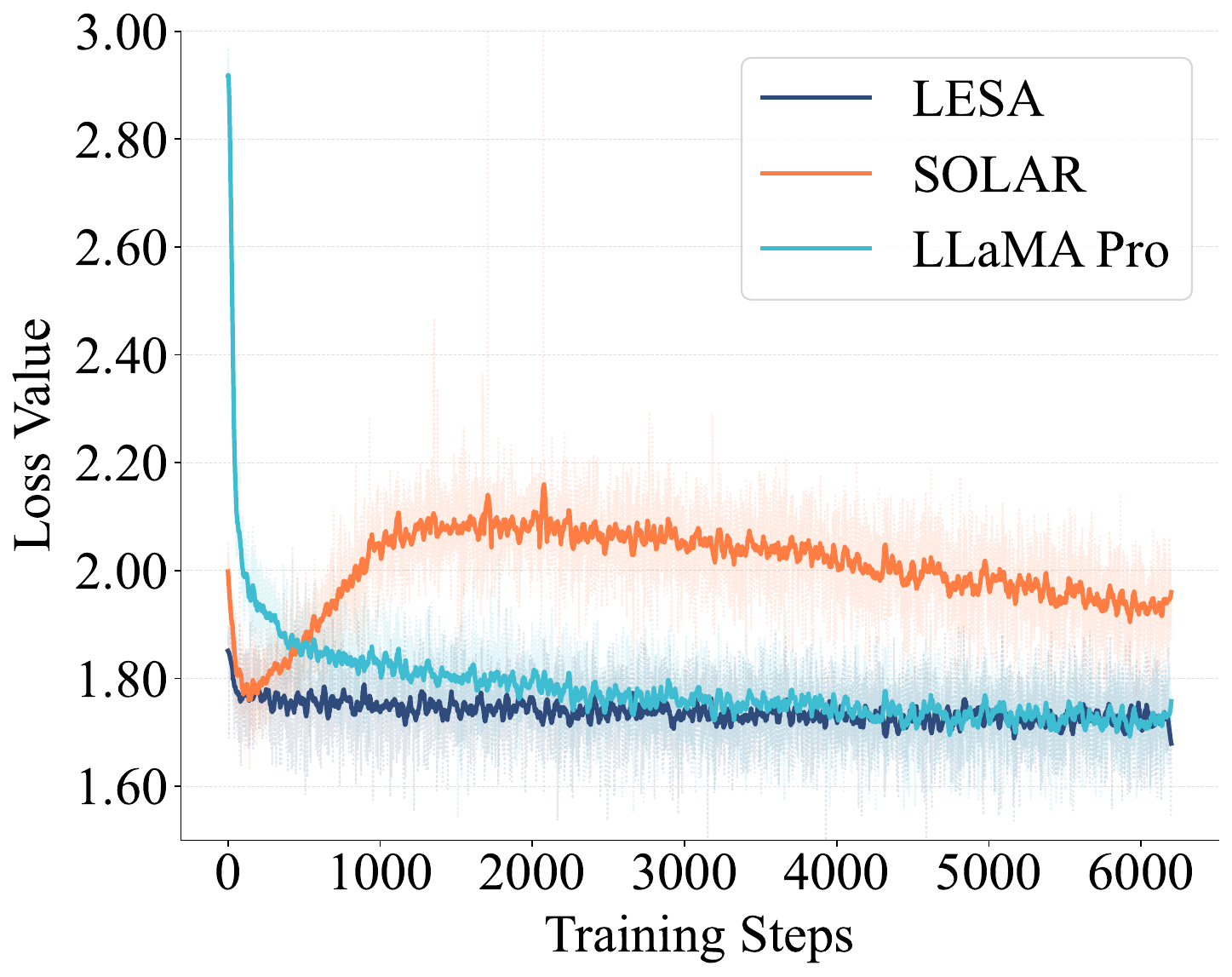}
    \caption{The continual pre-training loss curves of models expanded by different methods. \textbf{LESA} starts with a lower initial loss, indicating a better initialization. It stabilizes after 2k steps, reaching the same convergence level as LLaMA Pro after 5k steps, and converges much faster than SOLAR, achieving the same loss with less than half the training cost.}
    \label{fig:loss_curve}
\end{figure}

\begin{table}[!tp]
\small
\setlength\tabcolsep{6pt}
    \centering
    \begin{tabular}{cccc}
    \toprule
    \textbf{Model} & Pro & SOLAR & LESA \\
    \midrule
    \textbf{Time} & 56.4h\textcolor{mydarkgreen}{(124\%)} & 75.6h\textcolor{mydarkgreen}{(166\%)} & \textbf{45.6h} \\
    \bottomrule
    \end{tabular}
    \caption{Training time of continual pre-training. When trained on the same dataset, the baselines require 124\% and 166\% of the training time compared to our method.}
    \label{tab:training_time}
\end{table}

For the models after SFT, which have gained instruction-following capabilities, we use the generation mode of OpenCompass for evaluation. We conduct evaluations on ARC~\cite{clark2018think}, TruthfulQA~\cite{lin2021truthfulqa}, GSM8K~\cite{cobbe2021training}, HellaSwag~\cite{zellers2019hellaswag}, and MMLU~\cite{hendryckstest2021}.

\begin{table*}[!htp]
\setlength\tabcolsep{0.8pt} 
\small
    \centering
    \resizebox{0.97\linewidth}{!}{
    \begin{tabular}{c|c|c|ccc|cc|cc|cc|cc}
    \toprule
    \multirow{2}{*}{\textbf{Model}} & \multirow{2}{*}{\textbf{PPL}} & \multirow{2}{*}{\textbf{Average}} & \multicolumn{3}{c|}{\textbf{Reasoning}} & \multicolumn{2}{c|}{\textbf{Language}} & \multicolumn{2}{c|}{\textbf{Knowledge}} & \multicolumn{2}{c|}{\textbf{Examination}} & \multicolumn{2}{c}{\textbf{Understanding}} \\
    & &  & CMNLI & HeSw & PIQA & CHID & Wino & CSQA & BoolQ & MMLU & CMMLU &  Race$_{\text{H}}$ & Race$_{\text{M}}$ \\
    \midrule
    Pro-3k & 6.06 & 60.89\textcolor{mydarkgreen}{(-3.41)} & \textbf{32.99} & 69.14 & 79.21 & 67.33 & 55.79 & 69.19 & 68.10 & 45.53 & 49.66 & 65.32 & 67.55 \\
    \cmidrule{1-14}
    Pro-6k & 5.44 & 62.67\textcolor{mydarkgreen}{(-1.63)} & 32.97 & 70.15 & 78.94 & 69.80 & 55.09 & 69.04 & 66.79 & 64.31 & 50.30 & 62.66 & 69.36 \\
    \cmidrule{1-14}
    SOLAR-3k & 9.82 & 45.34\textcolor{mydarkgreen}{(-18.96)} & 32.97 & 61.11 & 73.23 & 51.49 & 53.68 & 54.55 & 52.94 & 34.98 & 28.97 & 27.36 & 27.44 \\
    \cmidrule{1-14}
    SOLAR-6k & 8.09 & 47.86\textcolor{mydarkgreen}{(-16.44)} & 32.98 & 62.36 & 74.65 & 48.02 & 54.38 & 59.54 & 61.53 & 43.74 & 28.68 & 31.76 & 28.83 \\
    \cmidrule{1-14}
    LESA-3k & 5.27 & 64.11\textcolor{mydarkgreen}{(-0.19)}& \textbf{32.99} & 71.18 & 79.65 & 72.77 & \textbf{57.89} & \textbf{69.78} & \textbf{70.46} & 66.66 & 50.91 & 65.32 & 67.55    \\
    \cmidrule{1-14}
    LESA-6k & \textbf{5.13} & \textbf{64.30} & \textbf{32.99} & \textbf{71.51} & \textbf{79.92} & \textbf{73.30} & 57.54 & 69.21 & 69.94 & \textbf{66.67} & \textbf{51.00} & \textbf{65.72} & \textbf{69.50}   \\
    \bottomrule
    \end{tabular}}
    \caption{We evaluate the performance of models after expanding Llama3-8B from 32 layers to 48 layers (11.5B parameters) using different baseline methods, followed by continual pre-training. Pro (LLaMA Pro~\cite{wu2024llama}) and SOLAR~\cite{kim2023solar} are two strong baselines for model depth scaling-up. We evaluate the model performance at two stages: after training with half the data (3k steps) and after training with the full data (6k steps).}
    \label{tab:main_res}
\end{table*}

\begin{table*}[!htbp]
\setlength\tabcolsep{8.0pt}
\small
    \centering
    \resizebox{0.97\linewidth}{!}{
    \begin{tabular}{cccccccc}
    \toprule
     Model & \textbf{Average} & \textbf{ARC-e} & \textbf{ARC-c} & \textbf{TruthfulQA} & \textbf{GSM8K} & \textbf{HellaSwag} & \textbf{MMLU} \\
    \midrule
    Pro-SFT & 24.38\textcolor{mydarkgreen}{(77\%)} & 28.92 & 23.73 & 21.91 & 21.95 & 25.44 & \textbf{24.33} \\
    \midrule
    SOLAR-SFT & 26.47\textcolor{mydarkgreen}{(84\%)} & 37.10 & 24.25 & 19.34 & 33.45 & 25.16 & 19.52 \\
    \midrule
    LESA-SFT & \textbf{31.57(100\%)} & \textbf{42.86} & \textbf{32.54} & \textbf{22.28} & \textbf{37.14} & \textbf{32.09} & 22.49 \\
    \bottomrule
    \end{tabular}}
    \caption{After continual pre-training and subsequent SFT, the model expanded with LESA still achieves better task performance, with baselines scoring less than 85\% of our model's average score.}
    \label{tab:sft_res}
\end{table*}

\subsection{Results}
We first present the training loss curves of the three models in Figure~\ref{fig:loss_curve}. From the figure, we observe that our method starts with a lower initial loss compared to the baselines, indicating a better initialization checkpoint. Throughout training, our model's loss consistently remains the lowest. SOLAR even fails to converge to a low loss level even after training on the dataset. Although LLaMA Pro's loss approaches ours after 5k steps, by the end, the model expanded using our method still has the lowest loss. Additionally, our method's loss stabilizes after 2k steps, while LLaMA Pro reaches a similar convergence level only after 5k steps. This demonstrates that models expanded using LESA achieve the same loss convergence with less than half the training cost.

We list the time taken to train on the full dataset in Table~\ref{tab:training_time} and find that the time taken by \LESA is significantly shorter. It is worth noting that the training of $\mathcal{G_W}$ in \LESA is very fast, taking less than 5 minutes, making its cost nearly negligible compared to the overhead of continual pre-training.

For model performance after continual pre-training, we present the results on various benchmarks in Table~\ref{tab:main_res}. It can be inferred that the performance of models expanded with \LESA consistently outperforms the baselines in all categories. Specifically, LESA-6k (6k steps) achieves the highest performance across all tasks and PPL. Even with only half of the data used for continual pre-training (3k steps), the models expanded using LESA outperform the baselines trained on the full dataset (6k steps). We also present the results in Table~\ref{tab:sft_res} for models trained on the full dataset and then fine-tuned with SFT, showing performance across different tasks. The results still indicate that the models expanded using LESA achieve the best performance.

The above analysis proves that \LESA effectively inherits the original model's parameters, enabling better initialization, faster continual training, and enhanced model performance.

\begin{table}[!tbp]
\small
    \centering
    \begin{tabular}{ccccc}
    \toprule
     Model & \textbf{CSQA} & \textbf{BoolQ} & \textbf{TriviaQA} & \textbf{NQ} \\
    \midrule
    Pro-3k & 69.19 & 68.10 & 62.50 & 24.82\\
    Pro-6k & 69.04 & 66.79 & 63.68 & 26.73\\\midrule
    SOLAR-3k & 54.55 & 52.94 & 43.06 & 13.57\\
    SOLAR-6k & 59.54 & 61.53 & 47.72 & 16.40\\\midrule
    LESA-3k & \textbf{69.78} & \textbf{70.46} & \textbf{67.15} & 23.30 \\
    LESA-6k & 69.21 & 69.94 & 67.05 & \textbf{26.76} \\
    \bottomrule
    \end{tabular}
    \caption{The scores of different models on knowledge-related tasks after continual pre-training. LESA consistently performs better overall.}
    \label{tab:know_res}
\end{table}

\subsection{Evaluation on Knowledge-Related Tasks}
Previous studies, such as LLaMA Pro, highlight that a key advantage of model expansion is the ability to inherit knowledge from the original model. We focus on evaluating performance in knowledge-related tasks. In addition to the main results, we further evaluate performance on two additional knowledge tasks: TriviaQA~\cite{JoshiTriviaQA2017} and NQ~\cite{kwiatkowski2019natural}. The results in Table~\ref{tab:know_res} show that \LESA outperforms previous approaches on all knowledge tasks.

\section{Ablation Study}
\subsection{Evaluation across Different Model Families}

We also aim to explore whether LESA is effective across different model sizes and families. Specifically, we select several current mainstream model families Llama3, Qwen2.5, Mistral~\cite{mistral3-small} and use LESA to expand the final layers of the models, increasing their layer count by 1.5x of the original. We use SOLAR initialization as the baseline. Since their method only applies to 32-layer models by concatenating the first 24 and last 24 layers, we adapt it for models with different layer counts. We concatenate the first and last $n$ layers to create a model with 1.5 times the original layers and measure initialization performance using PPL. The results are shown in Table~\ref{tab:more_model}.

\begin{table}[!htbp]
\small
    \centering
    \begin{tabular}{cccc}
    \toprule
     \textbf{Model} & \textbf{Original} & +\LESA & \textbf{+SOLAR} \\
    \midrule
    Llama3-8B & 5.20 & 6.35 & 7.81\\
    Llama3-70B & 1.98 & 2.62 & 4.21 \\\midrule
    Qwen2.5-1.5B & 9.30 & 10.52 & 11.75 \\
    Qwen2.5-7B & 6.03 & 7.04 & 7.99 \\
    Qwen2.5-32B & 3.78 & 5.67 & INF \\\midrule
    Mistral-Small-24B & 4.43 & 5.17 & 6.51 \\
    \bottomrule
    \end{tabular}
    \caption{PPL of LESA and SOLAR during 1.5x layer expansion initialization for different models, along with the PPL of the original models.}
    \label{tab:more_model}
\end{table}

The results show that LESA outperforms SOLAR in initialization performance. Unlike SOLAR, which experiences a PPL explosion on Qwen2.5-32B, LESA remains stable, highlighting the superiority of LESA's predicted parameters over SOLAR's heuristic-based expansion.

\subsection{Analysis of $\mathcal{G_W}$'s Ability}
We investigate whether $\mathcal{G_W}$ can predict intermediate layers between adjacent layers accurately, demonstrating this through loss changes.

Due to the limited number of samples available for training $\mathcal{G_W}$ on individual LLM layers, which makes it difficult to separate a test set and increases the risk of overfitting, we select several models: Llama3-8B, and fine-tuned versions of it, including Llama3-8B-Lexi-Uncensored~\cite{OrengutengLlama38BLexiUncensored}, Meta-Llama3-8B-Instruct, Llama-3-Smaug-8B~\cite{pal2024smaug}, and Llama3-8B-Chinese-Chat~\cite{shenzhi_wang_2024}. Following the procedure outlined in Section~\ref{sec:settings}, we sequentially select three consecutive layers as samples, resulting in a total of 150 samples. We use 120 samples for training and 30 for testing. The hyperparameters for training are set consistent with those used in the main experiment.

\begin{table}[!tbp]
\centering
\small
\begin{tabular}{cccc}
\hline
Matrix & \textbf{Random Loss} & \textbf{Training Loss} & \textbf{Test Loss} \\
\hline
down\_proj & 5.7 & 0.0005 & 0.0004 \\
up\_proj & 0.055 & 0.015 & 0.015 \\
gate\_proj & 0.056 & 0.015 & 0.015 \\
q\_proj & 0.153 & 0.016 & 0.016 \\
v\_proj & 0.545 & 0.017 & 0.016 \\
o\_proj & 0.147 & 0.016 & 0.015 \\
k\_proj & 0.6 & 0.016 & 0.016 \\
\hline
\end{tabular}
\caption{Loss values for different matrices during training and testing. All values are multiplied by $10^{4}$ for convenience.}
\label{tab:train_test_loss_values}
\end{table}

We present the loss values of $\mathcal{G_W}$ on both the training and test sets after training in Table~\ref{tab:train_test_loss_values}. For comparison, we also show the loss on the training set after random initialization. The results demonstrate that $\mathcal{G_W}$ significantly reduces the loss on the training set after training, typically lowering it to below 10\% of the random initialization loss. Moreover, the loss on the test set remains at the same level as the training set loss, indicating that $\mathcal{G_W}$ effectively learns the underlying patterns of the model parameters.

\begin{table}[!tbp]
\small
    \centering
    \begin{tabular}{cccc}
    \toprule
     \textbf{Method} & \textbf{Pro} & \textbf{SOLAR} & \LESA \\
    \midrule
    HumanEval & 10.98 & 2.44 & \textbf{25.00} \\
    MBPP & 21.69 & 13.93 & \textbf{28.60} \\
    \bottomrule
    \end{tabular}
    \caption{The results of Llama3-8B after expansion with different methods, pre-trained on the BigCode dataset, on two code benchmarks. The results show that LESA consistently performs better.}
    \label{tab:code_res}
\end{table}

\subsection{Single-Domain Pre-training}
In addition to general-domain pre-training experiments, we explore whether models expanded using our method show greater potential for continual pre-training in a single-domain setting. We conduct experiments in the code domain, using a subset of BigCode~\cite{Kocetkov2022TheStack}, one of the largest code pre-training datasets, while keeping other settings unchanged. Each model is trained for 40-60 hours and then evaluated on the HumanEval~\cite{chen2021codex} and MBPP~\cite{austin2021program} benchmarks. Table~\ref{tab:code_res} shows that after continual pre-training on the same code dataset, models expanded using our method outperform previous approaches, demonstrating its effectiveness in single-domain pre-training.

\subsection{Impact of SVD}
We observe inter-layer patterns of matrices in the SVD space, as shown in Figure~\ref{fig:gate_pattern}, which inspires us to train $\mathcal{G_W}$ in the SVD space for prediction. We also explore whether $\mathcal{G_W}$ can still predict effective matrices for layer expansion without SVD.

\begin{figure}[!tp]
    \centering    \includegraphics[width=0.98\linewidth,scale=1.00]{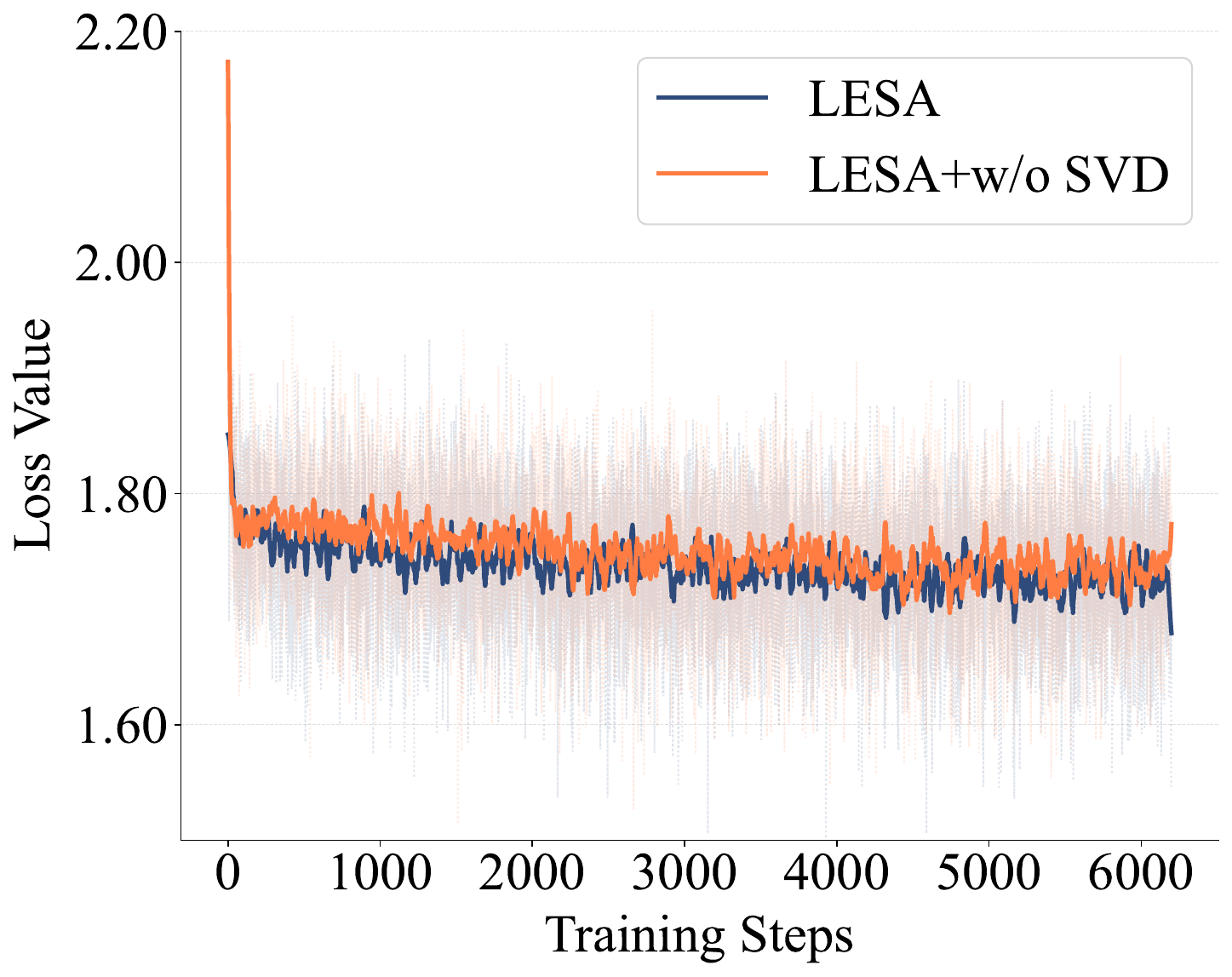}
    \caption{The continual pre-training loss curves without SVD, compared to the main experiment, show that with SVD, the model's initial loss and the final converged loss are both slightly lower.}
    \label{fig:loss_ab_svd}
\end{figure}

\begin{table}[!tbp]
\setlength\tabcolsep{0.8pt} 
\small
    \centering
    \begin{tabular}{ccccc}
    \toprule
     Model & \textbf{PIQA} & \textbf{BoolQ} & \textbf{HeSw} & \textbf{Wino} \\
    \midrule
    LESA-6k & 79.92 & 69.94 & 71.51 & 57.54\\
    \quad- SVD & 79.54\textcolor{mydarkgreen}{(-0.38)} & 68.81\textcolor{mydarkgreen}{(-0.13)} & 70.44\textcolor{mydarkgreen}{(-1.07)} & 57.29\textcolor{mydarkgreen}{(-0.25)}  \\\midrule
    Pro-6k & 78.94 & 66.79 & 70.15 & 55.09 \\
    \bottomrule
    \end{tabular}
    \caption{Without SVD, performance on several tasks is lower than with SVD, but still surpasses LLaMA Pro.}
    \label{tab:wo_svd_res}
\end{table}

We conduct an ablation study where we remove the SVD decomposition step while keeping other aspects of the method unchanged. Instead, we directly input the matrices to train $\mathcal{G_W}$, which predicts the parameters to be inserted between adjacent layers. We conduct experiments on Llama3-8B, expanding it to 48 layers and performing pre-training with the same data and hyper-parameters as in the main experiment. The loss curves with/without SVD are shown in Figure~\ref{fig:loss_ab_svd}. Without SVD, the model performs worse, with higher loss in the early stages and an average loss of 0.03 higher than with SVD after 3k steps. Thus, the addition of SVD is beneficial. We evaluate the models on several tasks, as shown in Table~\ref{tab:wo_svd_res}. The results show that while the model expanded without SVD performs slightly worse, it still outperforms the LLaMA Pro baseline. This demonstrates the effectiveness of LESA, with SVD further enhancing performance.

\begin{figure}[!tp]
    \centering    \includegraphics[width=0.98\linewidth,scale=1.00]{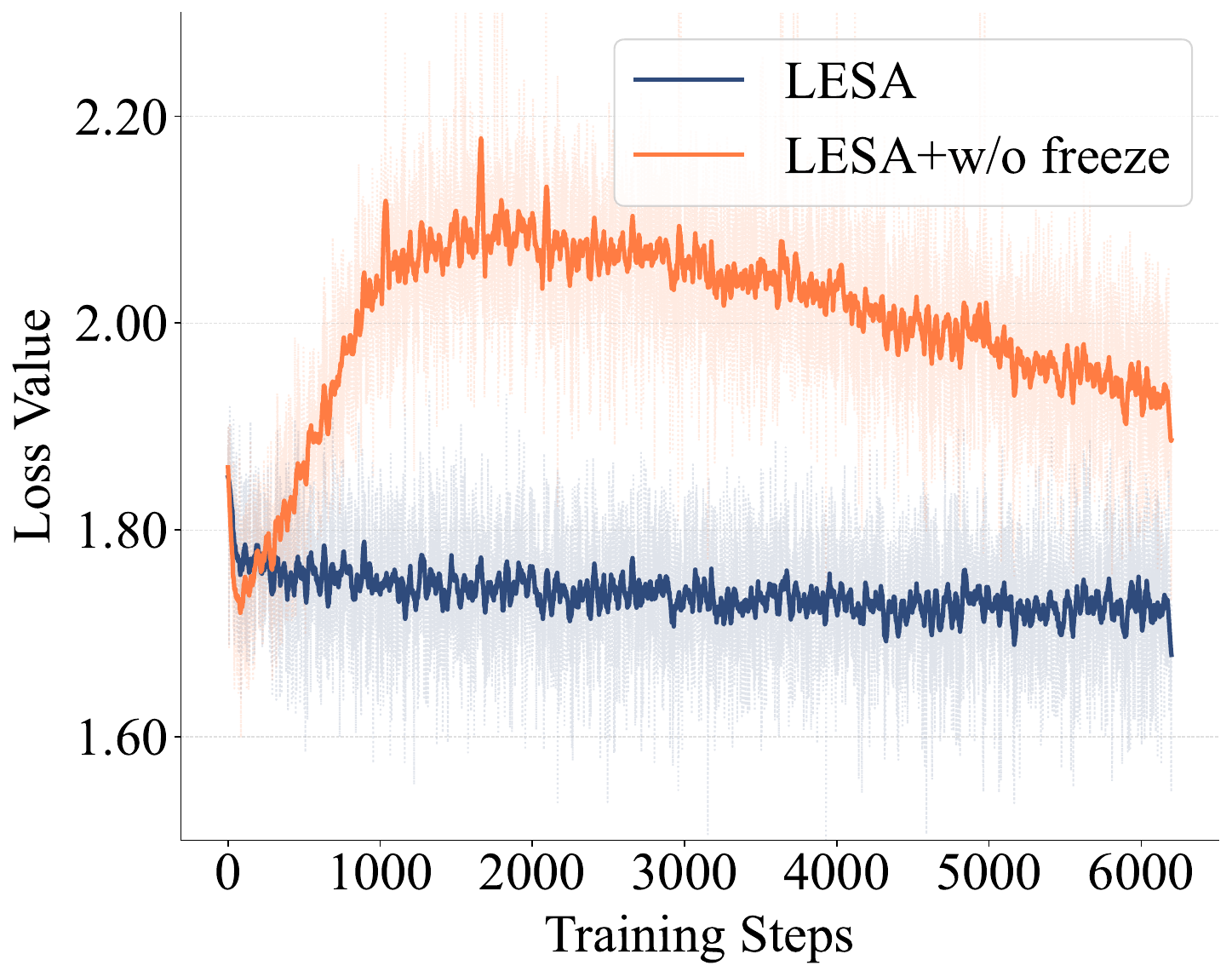}
    \caption{The training curves for LESA and LESA without freezing layers. When not freezing, the loss fluctuates and converges more slowly.}
    \label{fig:loss_curve_freeze}
\end{figure}

\subsection{Impact of Freezing Layers during Continual Pre-training}
Following LLaMA Pro, we train only the newly expanded layers during continual pre-training. We also explore full parameter fine-tuning without freezing any layers. Compared to the main experiment, we directly fine-tune all parameters while keeping the training data and hyperparameters consistent. The loss curves are shown in Figure~\ref{fig:loss_curve_freeze}. The figure shows that without freezing layers, loss converges much slower, with fluctuations in the curve. This suggests that, similar to LLaMA Pro, freezing the original parameters is essential for faster and better loss convergence.

More experiments on hyper-parameter settings, loss design, and the effectiveness on MoE model can be found in Appendix~\ref{app:more_exp}.

\section{Conclusion}
In this paper, we introduce \LESA, a novel approach for depth scaling-up of LLMs that overcomes the limitations of current heuristic-based methods. Using SVD and a neural network, \LESA predicts intermediate layer parameters, resulting in improved model initialization and faster convergence during continual pre-training. Extensive experiments show that \LESA outperforms existing baselines, delivering superior performance with lower computational costs. Furthermore, \LESA is effective across various model sizes, families, and domain-specific tasks, offering a promising solution for scaling LLMs efficiently. Our discovery of inter-layer patterns also provides new insights for future model design and training.

\section*{Limitations}
This work does not yet consider scaling the model to sizes larger than three times the parameters. Based on current model design practices, when increasing the number of layers significantly, it is typically necessary to expand the matrix size of each layer as well, which requires width scaling-up. We plan to explore this in future work.

Although we have conducted a preliminary exploration of LESA on MoE model, the research is still limited by the challenges of constructing routers for the predicted layers and the current large size of MoE models. Further investigation into MoE models is needed, and we consider this as future work.

\bibliography{custom}

\begin{thebibliography}{72}
\providecommand{\natexlab}[1]{#1}

\bibitem[{qwe(2024)}]{qwen2}
 2024.
\newblock Qwen2 technical report.

\bibitem[{Achiam et~al.(2023)Achiam, Adler, Agarwal, Ahmad, Akkaya, Aleman, Almeida, Altenschmidt, Altman, Anadkat et~al.}]{achiam2023gpt}
Josh Achiam, Steven Adler, Sandhini Agarwal, Lama Ahmad, Ilge Akkaya, Florencia~Leoni Aleman, Diogo Almeida, Janko Altenschmidt, Sam Altman, Shyamal Anadkat, et~al. 2023.
\newblock Gpt-4 technical report.
\newblock \emph{arXiv preprint arXiv:2303.08774}.

\bibitem[{Agarap(2018)}]{agarap2018deep}
AF~Agarap. 2018.
\newblock Deep learning using rectified linear units (relu).
\newblock \emph{arXiv preprint arXiv:1803.08375}.

\bibitem[{Agarwal et~al.(2024)Agarwal, Awasthi, Kale, and Zhao}]{agarwal2024stacking}
Naman Agarwal, Pranjal Awasthi, Satyen Kale, and Eric Zhao. 2024.
\newblock Stacking as accelerated gradient descent.
\newblock \emph{arXiv preprint arXiv:2403.04978}.

\bibitem[{AI@Meta(2024)}]{llama3modelcard}
AI@Meta. 2024.
\newblock \href {https://github.com/meta-llama/llama3/blob/main/MODEL_CARD.md} {Llama 3 model card}.

\bibitem[{Almazrouei et~al.(2023)Almazrouei, Alobeidli, Alshamsi, Cappelli, Cojocaru, Debbah, Goffinet, Hesslow, Launay, Malartic et~al.}]{almazrouei2023falcon}
Ebtesam Almazrouei, Hamza Alobeidli, Abdulaziz Alshamsi, Alessandro Cappelli, Ruxandra Cojocaru, M{\'e}rouane Debbah, {\'E}tienne Goffinet, Daniel Hesslow, Julien Launay, Quentin Malartic, et~al. 2023.
\newblock The falcon series of open language models.
\newblock \emph{arXiv preprint arXiv:2311.16867}.

\bibitem[{Austin et~al.(2021)Austin, Odena, Nye, Bosma, Michalewski, Dohan, Jiang, Cai, Terry, Le, and Sutton}]{austin2021program}
Jacob Austin, Augustus Odena, Maxwell Nye, Maarten Bosma, Henryk Michalewski, David Dohan, Ellen Jiang, Carrie Cai, Michael Terry, Quoc Le, and Charles Sutton. 2021.
\newblock \href {https://arxiv.org/abs/2108.07732} {Program synthesis with large language models}.
\newblock \emph{Preprint}, arXiv:2108.07732.

\bibitem[{Bai et~al.(2023)Bai, Bai, Chu, Cui, Dang, Deng, Fan, Ge, Han, Huang et~al.}]{bai2023qwen}
Jinze Bai, Shuai Bai, Yunfei Chu, Zeyu Cui, Kai Dang, Xiaodong Deng, Yang Fan, Wenbin Ge, Yu~Han, Fei Huang, et~al. 2023.
\newblock Qwen technical report.
\newblock \emph{arXiv preprint arXiv:2309.16609}.

\bibitem[{Bi et~al.(2024)Bi, Chen, Chen, Chen, Dai, Deng, Ding, Dong, Du, Fu et~al.}]{bi2024deepseek}
Xiao Bi, Deli Chen, Guanting Chen, Shanhuang Chen, Damai Dai, Chengqi Deng, Honghui Ding, Kai Dong, Qiushi Du, Zhe Fu, et~al. 2024.
\newblock Deepseek llm: Scaling open-source language models with longtermism.
\newblock \emph{arXiv preprint arXiv:2401.02954}.

\bibitem[{Bisk et~al.(2019)Bisk, Zellers, Bras, Gao, and Choi}]{bisk2019piqa}
Yonatan Bisk, Rowan Zellers, Ronan~Le Bras, Jianfeng Gao, and Yejin Choi. 2019.
\newblock \href {https://arxiv.org/abs/1911.11641} {Piqa: Reasoning about physical commonsense in natural language}.
\newblock \emph{Preprint}, arXiv:1911.11641.

\bibitem[{Brown et~al.(2020)Brown, Mann, Ryder, Subbiah, Kaplan, Dhariwal, Neelakantan, Shyam, Sastry, Askell et~al.}]{brown2020language}
Tom Brown, Benjamin Mann, Nick Ryder, Melanie Subbiah, Jared~D Kaplan, Prafulla Dhariwal, Arvind Neelakantan, Pranav Shyam, Girish Sastry, Amanda Askell, et~al. 2020.
\newblock Language models are few-shot learners.
\newblock \emph{Advances in neural information processing systems}, 33:1877--1901.

\bibitem[{Cao et~al.(2024)Cao, Yang, and Zhao}]{cao2024head}
Zouying Cao, Yifei Yang, and Hai Zhao. 2024.
\newblock Head-wise shareable attention for large language models.
\newblock \emph{arXiv preprint arXiv:2402.11819}.

\bibitem[{Chang et~al.(2017)Chang, Meng, Haber, Tung, and Begert}]{chang2017multi}
Bo~Chang, Lili Meng, Eldad Haber, Frederick Tung, and David Begert. 2017.
\newblock Multi-level residual networks from dynamical systems view.
\newblock \emph{arXiv preprint arXiv:1710.10348}.

\bibitem[{Chen et~al.(2021{\natexlab{a}})Chen, Yin, Shang, Jiang, Qin, Wang, Wang, Chen, Liu, and Liu}]{chen2021bert2bert}
Cheng Chen, Yichun Yin, Lifeng Shang, Xin Jiang, Yujia Qin, Fengyu Wang, Zhi Wang, Xiao Chen, Zhiyuan Liu, and Qun Liu. 2021{\natexlab{a}}.
\newblock bert2bert: Towards reusable pretrained language models.
\newblock \emph{arXiv preprint arXiv:2110.07143}.

\bibitem[{Chen et~al.(2021{\natexlab{b}})Chen, Tworek, Jun, Yuan, de~Oliveira~Pinto, Kaplan, Edwards, Burda, Joseph, Brockman, Ray, Puri, Krueger, Petrov, Khlaaf, Sastry, Mishkin, Chan, Gray, Ryder, Pavlov, Power, Kaiser, Bavarian, Winter, Tillet, Such, Cummings, Plappert, Chantzis, Barnes, Herbert-Voss, Guss, Nichol, Paino, Tezak, Tang, Babuschkin, Balaji, Jain, Saunders, Hesse, Carr, Leike, Achiam, Misra, Morikawa, Radford, Knight, Brundage, Murati, Mayer, Welinder, McGrew, Amodei, McCandlish, Sutskever, and Zaremba}]{chen2021codex}
Mark Chen, Jerry Tworek, Heewoo Jun, Qiming Yuan, Henrique~Ponde de~Oliveira~Pinto, Jared Kaplan, Harri Edwards, Yuri Burda, Nicholas Joseph, Greg Brockman, Alex Ray, Raul Puri, Gretchen Krueger, Michael Petrov, Heidy Khlaaf, Girish Sastry, Pamela Mishkin, Brooke Chan, Scott Gray, Nick Ryder, Mikhail Pavlov, Alethea Power, Lukasz Kaiser, Mohammad Bavarian, Clemens Winter, Philippe Tillet, Felipe~Petroski Such, Dave Cummings, Matthias Plappert, Fotios Chantzis, Elizabeth Barnes, Ariel Herbert-Voss, William~Hebgen Guss, Alex Nichol, Alex Paino, Nikolas Tezak, Jie Tang, Igor Babuschkin, Suchir Balaji, Shantanu Jain, William Saunders, Christopher Hesse, Andrew~N. Carr, Jan Leike, Josh Achiam, Vedant Misra, Evan Morikawa, Alec Radford, Matthew Knight, Miles Brundage, Mira Murati, Katie Mayer, Peter Welinder, Bob McGrew, Dario Amodei, Sam McCandlish, Ilya Sutskever, and Wojciech Zaremba. 2021{\natexlab{b}}.
\newblock \href {https://arxiv.org/abs/2107.03374} {Evaluating large language models trained on code}.

\bibitem[{Chen et~al.(2015)Chen, Goodfellow, and Shlens}]{chen2015net2net}
Tianqi Chen, Ian Goodfellow, and Jonathon Shlens. 2015.
\newblock Net2net: Accelerating learning via knowledge transfer.
\newblock \emph{arXiv preprint arXiv:1511.05641}.

\bibitem[{Clark et~al.(2019)Clark, Lee, Chang, Kwiatkowski, Collins, and Toutanova}]{clark2019boolq}
Christopher Clark, Kenton Lee, Ming-Wei Chang, Tom Kwiatkowski, Michael Collins, and Kristina Toutanova. 2019.
\newblock Boolq: Exploring the surprising difficulty of natural yes/no questions.
\newblock \emph{arXiv preprint arXiv:1905.10044}.

\bibitem[{Clark et~al.(2018)Clark, Cowhey, Etzioni, Khot, Sabharwal, Schoenick, and Tafjord}]{clark2018think}
Peter Clark, Isaac Cowhey, Oren Etzioni, Tushar Khot, Ashish Sabharwal, Carissa Schoenick, and Oyvind Tafjord. 2018.
\newblock Think you have solved question answering? try arc, the ai2 reasoning challenge.
\newblock \emph{arXiv preprint arXiv:1803.05457}.

\bibitem[{Cobbe et~al.(2021)Cobbe, Kosaraju, Bavarian, Chen, Jun, Kaiser, Plappert, Tworek, Hilton, Nakano et~al.}]{cobbe2021training}
Karl Cobbe, Vineet Kosaraju, Mohammad Bavarian, Mark Chen, Heewoo Jun, Lukasz Kaiser, Matthias Plappert, Jerry Tworek, Jacob Hilton, Reiichiro Nakano, et~al. 2021.
\newblock Training verifiers to solve math word problems.
\newblock \emph{arXiv preprint arXiv:2110.14168}.

\bibitem[{Contributors(2023)}]{2023opencompass}
OpenCompass Contributors. 2023.
\newblock Opencompass: A universal evaluation platform for foundation models.
\newblock \url{https://github.com/open-compass/opencompass}.

\bibitem[{DeepSeek-AI et~al.(2025)DeepSeek-AI, Guo, Yang, Zhang, Song, Zhang, Xu, Zhu, Ma, Wang, Bi, Zhang, Yu, Wu, Wu, Gou, Shao, Li, Gao, Liu, Xue, Wang, Wu, Feng, Lu, Zhao, Deng, Zhang, Ruan, Dai, Chen, Ji, Li, Lin, Dai, Luo, Hao, Chen, Li, Zhang, Bao, Xu, Wang, Ding, Xin, Gao, Qu, Li, Guo, Li, Wang, Chen, Yuan, Qiu, Li, Cai, Ni, Liang, Chen, Dong, Hu, Gao, Guan, Huang, Yu, Wang, Zhang, Zhao, Wang, Zhang, Xu, Xia, Zhang, Zhang, Tang, Li, Wang, Li, Tian, Huang, Zhang, Wang, Chen, Du, Ge, Zhang, Pan, Wang, Chen, Jin, Chen, Lu, Zhou, Chen, Ye, Wang, Yu, Zhou, Pan, Li, Zhou, Wu, Ye, Yun, Pei, Sun, Wang, Zeng, Zhao, Liu, Liang, Gao, Yu, Zhang, Xiao, An, Liu, Wang, Chen, Nie, Cheng, Liu, Xie, Liu, Yang, Li, Su, Lin, Li, Jin, Shen, Chen, Sun, Wang, Song, Zhou, Wang, Shan, Li, Wang, Wei, Zhang, Xu, Li, Zhao, Sun, Wang, Yu, Zhang, Shi, Xiong, He, Piao, Wang, Tan, Ma, Liu, Guo, Ou, Wang, Gong, Zou, He, Xiong, Luo, You, Liu, Zhou, Zhu, Xu, Huang, Li, Zheng, Zhu, Ma, Tang, Zha, Yan, Ren, Ren, Sha, Fu, Xu, Xie, Zhang,
  Hao, Ma, Yan, Wu, Gu, Zhu, Liu, Li, Xie, Song, Pan, Huang, Xu, Zhang, and Zhang}]{deepseekai2025deepseekr1incentivizingreasoningcapability}
DeepSeek-AI, Daya Guo, Dejian Yang, Haowei Zhang, Junxiao Song, Ruoyu Zhang, Runxin Xu, Qihao Zhu, Shirong Ma, Peiyi Wang, Xiao Bi, Xiaokang Zhang, Xingkai Yu, Yu~Wu, Z.~F. Wu, Zhibin Gou, Zhihong Shao, Zhuoshu Li, Ziyi Gao, Aixin Liu, Bing Xue, Bingxuan Wang, Bochao Wu, Bei Feng, Chengda Lu, Chenggang Zhao, Chengqi Deng, Chenyu Zhang, Chong Ruan, Damai Dai, Deli Chen, Dongjie Ji, Erhang Li, Fangyun Lin, Fucong Dai, Fuli Luo, Guangbo Hao, Guanting Chen, Guowei Li, H.~Zhang, Han Bao, Hanwei Xu, Haocheng Wang, Honghui Ding, Huajian Xin, Huazuo Gao, Hui Qu, Hui Li, Jianzhong Guo, Jiashi Li, Jiawei Wang, Jingchang Chen, Jingyang Yuan, Junjie Qiu, Junlong Li, J.~L. Cai, Jiaqi Ni, Jian Liang, Jin Chen, Kai Dong, Kai Hu, Kaige Gao, Kang Guan, Kexin Huang, Kuai Yu, Lean Wang, Lecong Zhang, Liang Zhao, Litong Wang, Liyue Zhang, Lei Xu, Leyi Xia, Mingchuan Zhang, Minghua Zhang, Minghui Tang, Meng Li, Miaojun Wang, Mingming Li, Ning Tian, Panpan Huang, Peng Zhang, Qiancheng Wang, Qinyu Chen, Qiushi Du, Ruiqi Ge, Ruisong
  Zhang, Ruizhe Pan, Runji Wang, R.~J. Chen, R.~L. Jin, Ruyi Chen, Shanghao Lu, Shangyan Zhou, Shanhuang Chen, Shengfeng Ye, Shiyu Wang, Shuiping Yu, Shunfeng Zhou, Shuting Pan, S.~S. Li, Shuang Zhou, Shaoqing Wu, Shengfeng Ye, Tao Yun, Tian Pei, Tianyu Sun, T.~Wang, Wangding Zeng, Wanjia Zhao, Wen Liu, Wenfeng Liang, Wenjun Gao, Wenqin Yu, Wentao Zhang, W.~L. Xiao, Wei An, Xiaodong Liu, Xiaohan Wang, Xiaokang Chen, Xiaotao Nie, Xin Cheng, Xin Liu, Xin Xie, Xingchao Liu, Xinyu Yang, Xinyuan Li, Xuecheng Su, Xuheng Lin, X.~Q. Li, Xiangyue Jin, Xiaojin Shen, Xiaosha Chen, Xiaowen Sun, Xiaoxiang Wang, Xinnan Song, Xinyi Zhou, Xianzu Wang, Xinxia Shan, Y.~K. Li, Y.~Q. Wang, Y.~X. Wei, Yang Zhang, Yanhong Xu, Yao Li, Yao Zhao, Yaofeng Sun, Yaohui Wang, Yi~Yu, Yichao Zhang, Yifan Shi, Yiliang Xiong, Ying He, Yishi Piao, Yisong Wang, Yixuan Tan, Yiyang Ma, Yiyuan Liu, Yongqiang Guo, Yuan Ou, Yuduan Wang, Yue Gong, Yuheng Zou, Yujia He, Yunfan Xiong, Yuxiang Luo, Yuxiang You, Yuxuan Liu, Yuyang Zhou, Y.~X. Zhu,
  Yanhong Xu, Yanping Huang, Yaohui Li, Yi~Zheng, Yuchen Zhu, Yunxian Ma, Ying Tang, Yukun Zha, Yuting Yan, Z.~Z. Ren, Zehui Ren, Zhangli Sha, Zhe Fu, Zhean Xu, Zhenda Xie, Zhengyan Zhang, Zhewen Hao, Zhicheng Ma, Zhigang Yan, Zhiyu Wu, Zihui Gu, Zijia Zhu, Zijun Liu, Zilin Li, Ziwei Xie, Ziyang Song, Zizheng Pan, Zhen Huang, Zhipeng Xu, Zhongyu Zhang, and Zhen Zhang. 2025.
\newblock \href {https://arxiv.org/abs/2501.12948} {Deepseek-r1: Incentivizing reasoning capability in llms via reinforcement learning}.
\newblock \emph{Preprint}, arXiv:2501.12948.

\bibitem[{Devlin et~al.(2018)Devlin, Chang, Lee, and Toutanova}]{devlin2018bert}
Jacob Devlin, Ming-Wei Chang, Kenton Lee, and Kristina Toutanova. 2018.
\newblock Bert: Pre-training of deep bidirectional transformers for language understanding.
\newblock \emph{arXiv preprint arXiv:1810.04805}.

\bibitem[{Dong et~al.(2020)Dong, Liu, Li, and Shang}]{dong2020towards}
Chengyu Dong, Liyuan Liu, Zichao Li, and Jingbo Shang. 2020.
\newblock Towards adaptive residual network training: A neural-ode perspective.
\newblock In \emph{International conference on machine learning}, pages 2616--2626. PMLR.

\bibitem[{Du et~al.(2024)Du, Luo, Qiu, Huang, Shen, Cheng, Guo, and Fu}]{Du2024Stacking}
Wenyu Du, Tongxu Luo, Zihan Qiu, Zeyu Huang, Yikang Shen, Reynold Cheng, Yike Guo, and Jie Fu. 2024.
\newblock Stacking your transformers: A closer look at model growth for efficient llm pre-training.
\newblock \emph{arXiv preprint arXiv:2405.15319}.

\bibitem[{Fayek et~al.(2020)Fayek, Cavedon, and Wu}]{fayek2020progressive}
Haytham~M Fayek, Lawrence Cavedon, and Hong~Ren Wu. 2020.
\newblock Progressive learning: A deep learning framework for continual learning.
\newblock \emph{Neural Networks}, 128:345--357.

\bibitem[{Gong et~al.(2019)Gong, He, Li, Qin, Wang, and Liu}]{gong2019efficient}
Linyuan Gong, Di~He, Zhuohan Li, Tao Qin, Liwei Wang, and Tieyan Liu. 2019.
\newblock Efficient training of bert by progressively stacking.
\newblock In \emph{International conference on machine learning}, pages 2337--2346. PMLR.

\bibitem[{Gu et~al.(2020)Gu, Liu, Yu, Li, Chen, and Han}]{gu2020transformer}
Xiaotao Gu, Liyuan Liu, Hongkun Yu, Jing Li, Chen Chen, and Jiawei Han. 2020.
\newblock On the transformer growth for progressive bert training.
\newblock \emph{arXiv preprint arXiv:2010.12562}.

\bibitem[{Hendrycks et~al.(2021)Hendrycks, Burns, Basart, Zou, Mazeika, Song, and Steinhardt}]{hendryckstest2021}
Dan Hendrycks, Collin Burns, Steven Basart, Andy Zou, Mantas Mazeika, Dawn Song, and Jacob Steinhardt. 2021.
\newblock Measuring massive multitask language understanding.
\newblock \emph{Proceedings of the International Conference on Learning Representations (ICLR)}.

\bibitem[{Jiang et~al.(2023)Jiang, Sablayrolles, Mensch, Bamford, Chaplot, Casas, Bressand, Lengyel, Lample, Saulnier et~al.}]{jiang2023mistral}
Albert~Q Jiang, Alexandre Sablayrolles, Arthur Mensch, Chris Bamford, Devendra~Singh Chaplot, Diego de~las Casas, Florian Bressand, Gianna Lengyel, Guillaume Lample, Lucile Saulnier, et~al. 2023.
\newblock Mistral 7b.
\newblock \emph{arXiv preprint arXiv:2310.06825}.

\bibitem[{Joshi et~al.(2017)Joshi, Choi, Weld, and Zettlemoyer}]{JoshiTriviaQA2017}
Mandar Joshi, Eunsol Choi, Daniel~S. Weld, and Luke Zettlemoyer. 2017.
\newblock Triviaqa: A large scale distantly supervised challenge dataset for reading comprehension.
\newblock In \emph{Proceedings of the 55th Annual Meeting of the Association for Computational Linguistics}, Vancouver, Canada. Association for Computational Linguistics.

\bibitem[{Kaplan et~al.(2020)Kaplan, McCandlish, Henighan, Brown, Chess, Child, Gray, Radford, Wu, and Amodei}]{kaplan2020scaling}
Jared Kaplan, Sam McCandlish, Tom Henighan, Tom~B Brown, Benjamin Chess, Rewon Child, Scott Gray, Alec Radford, Jeffrey Wu, and Dario Amodei. 2020.
\newblock Scaling laws for neural language models.
\newblock \emph{arXiv preprint arXiv:2001.08361}.

\bibitem[{Kim et~al.(2023)Kim, Park, Kim, Lee, Song, Kim, Kim, Kim, Lee, Kim et~al.}]{kim2023solar}
Dahyun Kim, Chanjun Park, Sanghoon Kim, Wonsung Lee, Wonho Song, Yunsu Kim, Hyeonwoo Kim, Yungi Kim, Hyeonju Lee, Jihoo Kim, et~al. 2023.
\newblock Solar 10.7 b: Scaling large language models with simple yet effective depth up-scaling.
\newblock \emph{arXiv preprint arXiv:2312.15166}.

\bibitem[{Kocetkov et~al.(2022)Kocetkov, Li, Ben~Allal, Li, Mou, Muñoz~Ferrandis, Jernite, Mitchell, Hughes, Wolf, Bahdanau, von Werra, and de~Vries}]{Kocetkov2022TheStack}
Denis Kocetkov, Raymond Li, Loubna Ben~Allal, Jia Li, Chenghao Mou, Carlos Muñoz~Ferrandis, Yacine Jernite, Margaret Mitchell, Sean Hughes, Thomas Wolf, Dzmitry Bahdanau, Leandro von Werra, and Harm de~Vries. 2022.
\newblock The stack: 3 tb of permissively licensed source code.
\newblock \emph{Preprint}.

\bibitem[{Kwiatkowski et~al.(2019)Kwiatkowski, Palomaki, Redfield, Collins, Parikh, Alberti, Epstein, Polosukhin, Devlin, Lee et~al.}]{kwiatkowski2019natural}
Tom Kwiatkowski, Jennimaria Palomaki, Olivia Redfield, Michael Collins, Ankur Parikh, Chris Alberti, Danielle Epstein, Illia Polosukhin, Jacob Devlin, Kenton Lee, et~al. 2019.
\newblock Natural questions: a benchmark for question answering research.
\newblock \emph{Transactions of the Association for Computational Linguistics}, 7:453--466.

\bibitem[{Lai et~al.(2017)Lai, Xie, Liu, Yang, and Hovy}]{lai2017race}
Guokun Lai, Qizhe Xie, Hanxiao Liu, Yiming Yang, and Eduard Hovy. 2017.
\newblock Race: Large-scale reading comprehension dataset from examinations.
\newblock \emph{arXiv preprint arXiv:1704.04683}.

\bibitem[{Li et~al.(2023)Li, Zhang, Koto, Yang, Zhao, Gong, Duan, and Baldwin}]{li2023cmmlu}
Haonan Li, Yixuan Zhang, Fajri Koto, Yifei Yang, Hai Zhao, Yeyun Gong, Nan Duan, and Timothy Baldwin. 2023.
\newblock \href {https://arxiv.org/abs/2306.09212} {Cmmlu: Measuring massive multitask language understanding in chinese}.
\newblock \emph{Preprint}, arXiv:2306.09212.

\bibitem[{Lin et~al.(2021)Lin, Hilton, and Evans}]{lin2021truthfulqa}
Stephanie Lin, Jacob Hilton, and Owain Evans. 2021.
\newblock Truthfulqa: Measuring how models mimic human falsehoods.
\newblock \emph{arXiv preprint arXiv:2109.07958}.

\bibitem[{Loshchilov(2017)}]{loshchilov2017decoupled}
I~Loshchilov. 2017.
\newblock Decoupled weight decay regularization.
\newblock \emph{arXiv preprint arXiv:1711.05101}.

\bibitem[{Lu et~al.(2024)Lu, Zhong, Wang, Guo, Zhu, Huang, Wang, Mi, Wang, Wang et~al.}]{lu2024yoda}
Jianqiao Lu, Wanjun Zhong, Yufei Wang, Zhijiang Guo, Qi~Zhu, Wenyong Huang, Yanlin Wang, Fei Mi, Baojun Wang, Yasheng Wang, et~al. 2024.
\newblock Yoda: Teacher-student progressive learning for language models.
\newblock \emph{arXiv preprint arXiv:2401.15670}.

\bibitem[{Men et~al.(2024)Men, Xu, Zhang, Wang, Lin, Lu, Han, and Chen}]{men2024shortgpt}
Xin Men, Mingyu Xu, Qingyu Zhang, Bingning Wang, Hongyu Lin, Yaojie Lu, Xianpei Han, and Weipeng Chen. 2024.
\newblock Shortgpt: Layers in large language models are more redundant than you expect.
\newblock \emph{arXiv preprint arXiv:2403.03853}.

\bibitem[{Mistral@AI(2025)}]{mistral3-small}
Mistral@AI. 2025.
\newblock \href {https://huggingface.co/mistralai/Mistral-Small-24B-Base-2501} {Mistral small 3}.

\bibitem[{Orenguteng(2024)}]{OrengutengLlama38BLexiUncensored}
Orenguteng. 2024.
\newblock \href {https://huggingface.co/Orenguteng/Llama-3-8B-Lexi-Uncensored} {Llama-3-8b-lexi-uncensored}.

\bibitem[{Pal et~al.(2024)Pal, Karkhanis, Dooley, Roberts, Naidu, and White}]{pal2024smaug}
Arka Pal, Deep Karkhanis, Samuel Dooley, Manley Roberts, Siddartha Naidu, and Colin White. 2024.
\newblock Smaug: Fixing failure modes of preference optimisation with dpo-positive.
\newblock \emph{arXiv preprint arXiv:2402.13228}.

\bibitem[{Pan et~al.(2024)Pan, Yuan, Yin, Shi, Xu, Zhang, Shang, Jiang, and Liu}]{pan2024preparing}
Yu~Pan, Ye~Yuan, Yichun Yin, Jiaxin Shi, Zenglin Xu, Ming Zhang, Lifeng Shang, Xin Jiang, and Qun Liu. 2024.
\newblock Preparing lessons for progressive training on language models.
\newblock In \emph{Proceedings of the AAAI Conference on Artificial Intelligence}, volume~38, pages 18860--18868.

\bibitem[{Parmar et~al.(2024)Parmar, Satheesh, Patwary, Shoeybi, and Catanzaro}]{parmar2024reuse}
Jupinder Parmar, Sanjev Satheesh, Mostofa Patwary, Mohammad Shoeybi, and Bryan Catanzaro. 2024.
\newblock Reuse, don't retrain: A recipe for continued pretraining of language models.
\newblock \emph{arXiv preprint arXiv:2407.07263}.

\bibitem[{Peng et~al.(2023)Peng, Li, He, Galley, and Gao}]{peng2023instruction}
Baolin Peng, Chunyuan Li, Pengcheng He, Michel Galley, and Jianfeng Gao. 2023.
\newblock Instruction tuning with gpt-4.
\newblock \emph{arXiv preprint arXiv:2304.03277}.

\bibitem[{Sakaguchi et~al.(2019)Sakaguchi, Bras, Bhagavatula, and Choi}]{sakaguchi2019winogrande}
Keisuke Sakaguchi, Ronan~Le Bras, Chandra Bhagavatula, and Yejin Choi. 2019.
\newblock Winogrande: An adversarial winograd schema challenge at scale.
\newblock \emph{arXiv preprint arXiv:1907.10641}.

\bibitem[{Samragh et~al.(2024)Samragh, Mirzadeh, Vahid, Faghri, Cho, Nabi, Naik, and Farajtabar}]{samragh2024scaling}
Mohammad Samragh, Iman Mirzadeh, Keivan~Alizadeh Vahid, Fartash Faghri, Minsik Cho, Moin Nabi, Devang Naik, and Mehrdad Farajtabar. 2024.
\newblock Scaling smart: Accelerating large language model pre-training with small model initialization.
\newblock \emph{arXiv preprint arXiv:2409.12903}.

\bibitem[{Stoica et~al.(2024)Stoica, Ramesh, Ecsedi, Choshen, and Hoffman}]{stoica2024model}
George Stoica, Pratik Ramesh, Boglarka Ecsedi, Leshem Choshen, and Judy Hoffman. 2024.
\newblock Model merging with svd to tie the knots.
\newblock \emph{arXiv preprint arXiv:2410.19735}.

\bibitem[{Talmor et~al.(2018)Talmor, Herzig, Lourie, and Berant}]{talmor2018commonsenseqa}
Alon Talmor, Jonathan Herzig, Nicholas Lourie, and Jonathan Berant. 2018.
\newblock Commonsenseqa: A question answering challenge targeting commonsense knowledge.
\newblock \emph{arXiv preprint arXiv:1811.00937}.

\bibitem[{Touvron et~al.(2023{\natexlab{a}})Touvron, Lavril, Izacard, Martinet, Lachaux, Lacroix, Rozi{\`e}re, Goyal, Hambro, Azhar et~al.}]{touvron2023llama}
Hugo Touvron, Thibaut Lavril, Gautier Izacard, Xavier Martinet, Marie-Anne Lachaux, Timoth{\'e}e Lacroix, Baptiste Rozi{\`e}re, Naman Goyal, Eric Hambro, Faisal Azhar, et~al. 2023{\natexlab{a}}.
\newblock Llama: Open and efficient foundation language models.
\newblock \emph{arXiv preprint arXiv:2302.13971}.

\bibitem[{Touvron et~al.(2023{\natexlab{b}})Touvron, Martin, Stone, Albert, Almahairi, Babaei, Bashlykov, Batra, Bhargava, Bhosale et~al.}]{touvron2023llama2}
Hugo Touvron, Louis Martin, Kevin Stone, Peter Albert, Amjad Almahairi, Yasmine Babaei, Nikolay Bashlykov, Soumya Batra, Prajjwal Bhargava, Shruti Bhosale, et~al. 2023{\natexlab{b}}.
\newblock Llama 2: Open foundation and fine-tuned chat models.
\newblock \emph{arXiv preprint arXiv:2307.09288}.

\bibitem[{Van~der Maaten and Hinton(2008)}]{van2008visualizing}
Laurens Van~der Maaten and Geoffrey Hinton. 2008.
\newblock Visualizing data using t-sne.
\newblock \emph{Journal of machine learning research}, 9(11).

\bibitem[{Vaswani et~al.(2017)Vaswani, Shazeer, Parmar, Uszkoreit, Jones, Gomez, Kaiser, and Polosukhin}]{vaswani2017attention}
Ashish Vaswani, Noam Shazeer, Niki Parmar, Jakob Uszkoreit, Llion Jones, Aidan~N Gomez, {\L}ukasz Kaiser, and Illia Polosukhin. 2017.
\newblock Attention is all you need.
\newblock \emph{Advances in neural information processing systems}, 30.

\bibitem[{Voita et~al.(2019{\natexlab{a}})Voita, Sennrich, and Titov}]{voita2019bottom}
Elena Voita, Rico Sennrich, and Ivan Titov. 2019{\natexlab{a}}.
\newblock The bottom-up evolution of representations in the transformer: A study with machine translation and language modeling objectives.
\newblock \emph{arXiv preprint arXiv:1909.01380}.

\bibitem[{Voita et~al.(2019{\natexlab{b}})Voita, Talbot, Moiseev, Sennrich, and Titov}]{voita2019analyzing}
Elena Voita, David Talbot, Fedor Moiseev, Rico Sennrich, and Ivan Titov. 2019{\natexlab{b}}.
\newblock Analyzing multi-head self-attention: Specialized heads do the heavy lifting, the rest can be pruned.
\newblock \emph{arXiv preprint arXiv:1905.09418}.

\bibitem[{Wang et~al.(2024{\natexlab{a}})Wang, Chen, Lin, Li, and Zhang}]{wang2024basis}
Jingcun Wang, Yu-Guang Chen, Ing-Chao Lin, Bing Li, and Grace~Li Zhang. 2024{\natexlab{a}}.
\newblock Basis sharing: Cross-layer parameter sharing for large language model compression.
\newblock \emph{arXiv preprint arXiv:2410.03765}.

\bibitem[{Wang et~al.(2023)Wang, Panda, Hennigen, Greengard, Karlinsky, Feris, Cox, Wang, and Kim}]{wang2023learning}
Peihao Wang, Rameswar Panda, Lucas~Torroba Hennigen, Philip Greengard, Leonid Karlinsky, Rogerio Feris, David~Daniel Cox, Zhangyang Wang, and Yoon Kim. 2023.
\newblock Learning to grow pretrained models for efficient transformer training.
\newblock \emph{arXiv preprint arXiv:2303.00980}.

\bibitem[{Wang et~al.(2024{\natexlab{b}})Wang, Zheng, Wang, Song, and Huang}]{shenzhi_wang_2024}
Shenzhi Wang, Yaowei Zheng, Guoyin Wang, Shiji Song, and Gao Huang. 2024{\natexlab{b}}.
\newblock \href {https://doi.org/10.57967/hf/2316} {Llama3-8b-chinese-chat (revision 6622a23)}.

\bibitem[{Wang et~al.(2024{\natexlab{c}})Wang, Zheng, Wan, and Zhang}]{wang2024svd}
Xin Wang, Yu~Zheng, Zhongwei Wan, and Mi~Zhang. 2024{\natexlab{c}}.
\newblock Svd-llm: Truncation-aware singular value decomposition for large language model compression.
\newblock \emph{arXiv preprint arXiv:2403.07378}.

\bibitem[{Wei et~al.(2016)Wei, Wang, Rui, and Chen}]{wei2016network}
Tao Wei, Changhu Wang, Yong Rui, and Chang~Wen Chen. 2016.
\newblock Network morphism.
\newblock In \emph{International conference on machine learning}, pages 564--572. PMLR.

\bibitem[{Wen et~al.(2020)Wen, Yan, Chen, and Li}]{wen2020autogrow}
Wei Wen, Feng Yan, Yiran Chen, and Hai Li. 2020.
\newblock Autogrow: Automatic layer growing in deep convolutional networks.
\newblock In \emph{Proceedings of the 26th ACM SIGKDD International Conference on Knowledge Discovery \& Data Mining}, pages 833--841.

\bibitem[{Wu et~al.(2024)Wu, Gan, Ge, Lu, Wang, Feng, Luo, and Shan}]{wu2024llama}
Chengyue Wu, Yukang Gan, Yixiao Ge, Zeyu Lu, Jiahao Wang, Ye~Feng, Ping Luo, and Ying Shan. 2024.
\newblock Llama pro: Progressive llama with block expansion.
\newblock \emph{arXiv preprint arXiv:2401.02415}.

\bibitem[{Xu et~al.(2020)Xu, Hu, Zhang, Li, Cao, Li, Xu, Sun, Yu, Yu et~al.}]{xu2020clue}
Liang Xu, Hai Hu, Xuanwei Zhang, Lu~Li, Chenjie Cao, Yudong Li, Yechen Xu, Kai Sun, Dian Yu, Cong Yu, et~al. 2020.
\newblock Clue: A chinese language understanding evaluation benchmark.
\newblock \emph{arXiv preprint arXiv:2004.05986}.

\bibitem[{Yang et~al.(2024{\natexlab{a}})Yang, Yang, Hui, Zheng, Yu, Zhou, Li, Li, Liu, Huang et~al.}]{yang2024qwen2}
An~Yang, Baosong Yang, Binyuan Hui, Bo~Zheng, Bowen Yu, Chang Zhou, Chengpeng Li, Chengyuan Li, Dayiheng Liu, Fei Huang, et~al. 2024{\natexlab{a}}.
\newblock Qwen2 technical report.
\newblock \emph{arXiv preprint arXiv:2407.10671}.

\bibitem[{Yang et~al.(2020)Yang, Wang, Yang, Li, He, and Zhang}]{yang2020progressively}
Cheng Yang, Shengnan Wang, Chao Yang, Yuechuan Li, Ru~He, and Jingqiao Zhang. 2020.
\newblock Progressively stacking 2.0: A multi-stage layerwise training method for bert training speedup.
\newblock \emph{arXiv preprint arXiv:2011.13635}.

\bibitem[{Yang et~al.(2024{\natexlab{b}})Yang, Cao, and Zhao}]{yang2024laco}
Yifei Yang, Zouying Cao, and Hai Zhao. 2024{\natexlab{b}}.
\newblock Laco: Large language model pruning via layer collapse.
\newblock \emph{arXiv preprint arXiv:2402.11187}.

\bibitem[{Yao et~al.(2023)Yao, Zhang, Li, and Wang}]{yao2023masked}
Yiqun Yao, Zheng Zhang, Jing Li, and Yequan Wang. 2023.
\newblock Masked structural growth for 2x faster language model pre-training.
\newblock \emph{arXiv preprint arXiv:2305.02869}.

\bibitem[{Zellers et~al.(2019)Zellers, Holtzman, Bisk, Farhadi, and Choi}]{zellers2019hellaswag}
Rowan Zellers, Ari Holtzman, Yonatan Bisk, Ali Farhadi, and Yejin Choi. 2019.
\newblock Hellaswag: Can a machine really finish your sentence?
\newblock \emph{arXiv preprint arXiv:1905.07830}.

\bibitem[{Zheng et~al.(2019)Zheng, Huang, and Sun}]{zheng-etal-2019-chid}
Chujie Zheng, Minlie Huang, and Aixin Sun. 2019.
\newblock {C}h{ID}: A large-scale {C}hinese {ID}iom dataset for cloze test.
\newblock In \emph{ACL}.

\bibitem[{Zheng et~al.(2024)Zheng, Zhang, Zhang, Ye, Luo, Feng, and Ma}]{zheng2024llamafactory}
Yaowei Zheng, Richong Zhang, Junhao Zhang, Yanhan Ye, Zheyan Luo, Zhangchi Feng, and Yongqiang Ma. 2024.
\newblock \href {http://arxiv.org/abs/2403.13372} {Llamafactory: Unified efficient fine-tuning of 100+ language models}.
\newblock In \emph{Proceedings of the 62nd Annual Meeting of the Association for Computational Linguistics (Volume 3: System Demonstrations)}, Bangkok, Thailand. Association for Computational Linguistics.

\bibitem[{Zhu et~al.(2024)Zhu, Qu, Dong, Ruan, Tong, He, and Cheng}]{llama-moe}
Tong Zhu, Xiaoye Qu, Daize Dong, Jiacheng Ruan, Jingqi Tong, Conghui He, and Yu~Cheng. 2024.
\newblock \href {https://arxiv.org/abs/2406.16554} {Llama-moe: Building mixture-of-experts from llama with continual pre-training}.
\newblock \emph{arXiv preprint arXiv:2406.16554}.

\end{thebibliography}

\appendix

\section{More Experiments}
\label{app:more_exp}
In this section, we provide additional experiments and analyses on hyperparameter settings, loss design, and the effectiveness on the MoE model.

\subsection{Impact of Layer Insertion Location}
Previous studies~\cite{yang2024laco,men2024shortgpt,cao2024head} suggest that LLMs are generally less sensitive to layers near the output end, which can be modified. Therefore, our main experiment focuses on expanding layers closer to the output end. We also aim to explore the performance of our method when expanding layers near the input end. Building on the main experiment, we change the range of the expanded layers from the original 15th to 31st layers to the 1st to 17th layers. We then compare the PPL on Wikipedia for the models after initialization, without further training.

\begin{table}[!htbp]
\small
    \centering
    \begin{tabular}{ccc}
    \toprule
    \textbf{Layer Interval} & \textbf{15-31} & \textbf{1-17} \\
    \midrule
    PPL & 6.35 & 57.32\\
    \bottomrule
    \end{tabular}
    \caption{The model's initialization performance is better when layers are inserted at the output than at the input end.}
    \label{tab:ins_lay}
\end{table}

The results in Table~\ref{tab:ins_lay} show that expanding layers near the input end results in poorer initialization performance than expanding near the output. This suggests that our method is more effective when layers are inserted closer to the output, aligning with previous findings.

\subsection{Ablation on Norm Loss}
We investigate whether it is possible to train $\mathcal{G_W}$ without adding the norm loss $\mathcal{L}_2$. Compared to the main experiment, we remove this loss and calculate the average norm of the matrices in the newly inserted layers predicted by $\mathcal{G_W}$.

\begin{table}[!htbp]
\small
    \centering
    \begin{tabular}{cccc}
    \toprule
      \textbf{Model} & \textbf{Llama3-8B} & \textbf{+LESA} & \textbf{+LESA+w/o $\mathcal{L}_2$} \\
    \midrule
    down\_proj & 80.88 & 80.70 & 13.18 \\
    up\_proj & 81.88 & 81.57 & 10.26 \\
    gate\_proj & 104.96 & 105.39 & 13.34\\
    q\_proj & 69.16 & 69.28 & 8.85 \\
    k\_proj & 52.44 & 51.01 & 7.32\\
    v\_proj & 19.88 & 18.98 & 2.91\\
    o\_proj & 40.25 & 40.49 & 5.29 \\
    \bottomrule
    \end{tabular}
    \caption{Without the norm loss $\mathcal{L}_2$, the norms of the matrices predicted by $\mathcal{G_W}$ are very small, leading to parameter degradation.}
    \label{tab:ab_norm}
\end{table}

As shown in Table~\ref{tab:ab_norm}, without $\mathcal{L}_2$, the predicted matrices have very small norms, causing their values to approach zero and leading to degeneration. However, with $\mathcal{L}_2$, the norms of the predicted matrices align with those of the original Llama3-8B matrices.

\subsection{Hyper-parameter Impact on Model Initialization}

In this section, we explore the impact of key hyper-parameters during the training of $\mathcal{G_W}$. We find that the number of epochs and learning rate affect the initialization performance of the model obtained through layer expansion. We also conduct experiments on Llama3-8B, varying the learning rate and epochs while keeping other hyper-parameters consistent with the main experiment.

\begin{table}[!htbp]
\small
    \centering
    \begin{tabular}{ccc}
    \toprule
    \textbf{Learning Rate} & \textbf{Epoch} & \textbf{PPL} \\
    \midrule
    1e-3 & 5 & 6.35 \\
    1e-4 & 5 & 102.42 \\
    5e-4 & 5 & 6.82 \\
    1e-4 & 10 & 39182.51 \\
    5e-4 & 10 & 6.94 \\
    \bottomrule
    \end{tabular}
    \caption{Ablation study on the hyperparameters during the training of $\mathcal{W_G}$.}
    \label{tab:hyper_study}
\end{table}

The results in Table~\ref{tab:hyper_study} show that adjusting the learning rate and epochs can sometimes cause the expanded model's PPL to explode during initialization. This may be due to the limited number of training samples generated from a single model, leading to training instability. However, after tuning the hyper-parameters a few times, we are able to achieve a good initialization performance, with PPL values typically ranging between 6 and 7.

Additionally, we find that the hidden-state size and the number of layers in $\mathcal{G_W}$ have no significant impact on the performance of the expanded model. The loss's $ \lambda$ only affects the matrix norm, but has minimal effect on the model's performance. Adjusting $\lambda$ to match the predicted matrix norm with that of the original model is sufficient.

\subsection{Effectiveness on MoE Model}
Recently, LLMs based on the Mixture-of-Experts (MoE) architecture have become increasingly popular. In this section, we explore the effectiveness of \LESA on such models. Due to the large size of current MoE models, such as DeepSeek-R1 with 671B parameters~\cite{deepseekai2025deepseekr1incentivizingreasoningcapability}, which cannot be loaded onto our server, we conduct experiments on the smaller LLaMA-MoE-3.0B~\cite{llama-moe}, which has 32 layers.

We use LESA to expand the model to 48 layers. However, a unique aspect of MoE models is that each layer has an MLP router, and we have not yet devised a method to generate routers for the newly added layers, since the router is highly dependent on the performance of each expert. Our current approach is to replicate the previous layer's router for the newly expanded layer. We use SOLAR as the baseline and then evaluate the PPL of the expanded model after initialization. The results are shown in Table~\ref{tab:moe_res}.

\begin{table}[!htbp]
\small
    \centering
    \begin{tabular}{cccc}
    \toprule
    \textbf{Model} & \textbf{+LLaMA-MoE-3.0B} & +\LESA & \textbf{+SOLAR}\\
    \midrule
    PPL & 7.70 & 1923.14 & 76.50 \\
    \bottomrule
    \end{tabular}
    \caption{The MoE model's initialization performance on PPL with different scaling-up methods.}
    \label{tab:moe_res}
\end{table}

The results show that LESA experiences a significant increase in PPL, which we attribute to the mismatch between the router and the expanded parameters. We will continue investigating this issue in future work. Meanwhile, SOLAR also performs poorly, increasing PPL by 10 times. This suggests that scaling-up methods for MoE models require further research.

\section{SVD-Based Patterns}\label{app:svd_pattern}
\begin{figure*}[!tp]
    \centering
    \includegraphics[width=0.98\linewidth,scale=1.00]{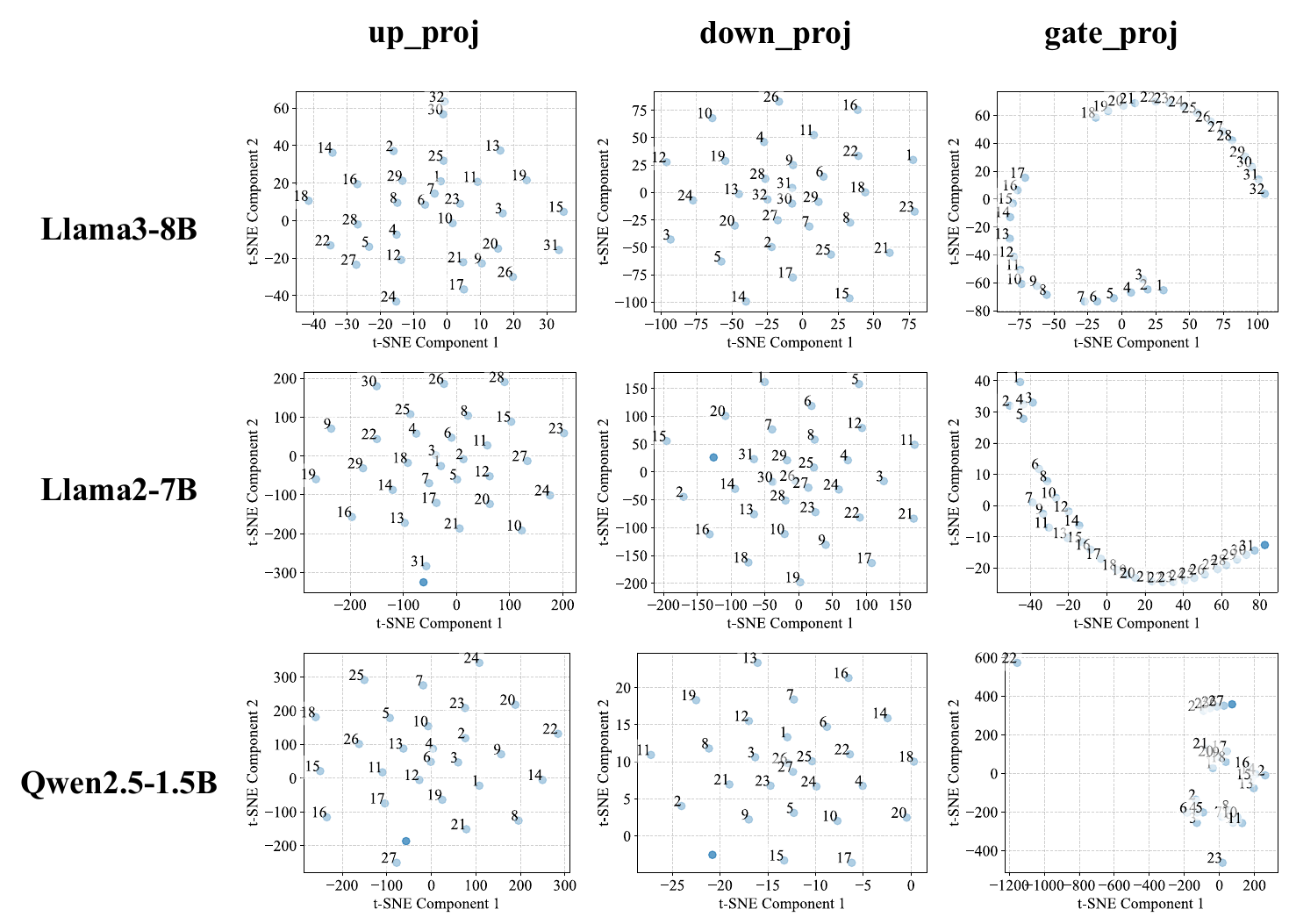}
    \caption{The gate\_proj parameter matrices in the MLP of different models exhibit clear patterns of continuity or clustering. This suggests that after applying SVD, the model's parameters may be learnable. The parameter distributions of other matrices appear more uniform in our visualizations.}
    \label{fig:mlp_pattern}
\end{figure*}

\begin{figure*}[!tp]
    \centering
    \includegraphics[width=0.98\linewidth,scale=1.00]{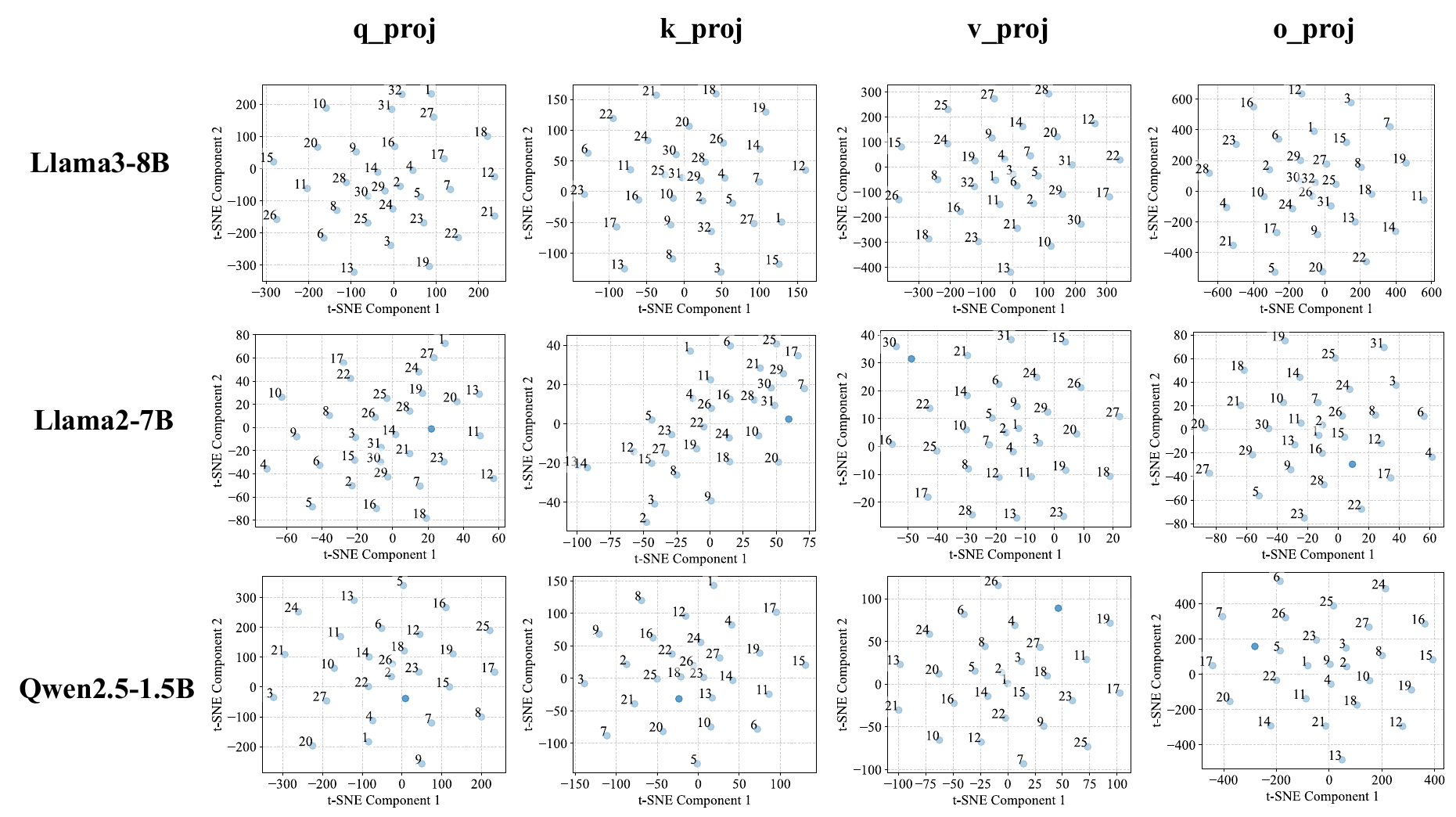}
    \caption{The parameter distributions of the matrices in the self-attention layers across different models appear relatively uniform in our visualizations.}
    \label{fig:self_atten_pattern}
\end{figure*}

We present the t-SNE visualizations of the top 1 singular values corresponding to the vectors of $V$, obtained after applying SVD decomposition to the matrices in the MLP and self-attention of different models, in Figure~\ref{fig:mlp_pattern} and Figure~\ref{fig:self_atten_pattern}, respectively.

\end{document}